\documentclass[sigconf]{acmart}
\AtBeginDocument{%
  }

\setcopyright{none}
\renewcommand\footnotetextcopyrightpermission[1]{}
\acmConference[Preprint]{Preprint. Under review}{}{}
\usepackage[table]{xcolor}
\usepackage{booktabs}
\usepackage{enumitem}
\usepackage{amsmath}
\usepackage{cleveref}
\usepackage{tabularx}
\usepackage{ragged2e} % For \RaggedRight alignment in X columns
\usepackage{multirow}
\usepackage{array}
\usepackage{pifont}
\usepackage[most]{tcolorbox}
\usepackage{algorithm}
\usepackage{algpseudocode}

\newcolumntype{C}[1]{>{\centering\arraybackslash}p{#1}}
\newcommand{\cmark}{\textcolor{green}{\ding{51}}} % ✓
\newcommand{\xmark}{\textcolor{red}{\ding{55}}} % ✗

\makeatletter
\patchcmd{\@mkauthors@iii}{\UrlBreakPenalty=10000}{\UrlBreakPenalty=100}{}{}
\patchcmd{\@mkauthors@iii}{\UrlBreakPenalty=10000}{\UrlBreakPenalty=100}{}{}
\makeatother

\begin{document}

%% Pack the 7 authors into 2 rows (4 + 3) instead of the default 3-per-row.
\settopmatter{authorsperrow=4}

\title{TREK: A Travel Reasoning and Evaluation Kit for LLM Agents in Complex Trip Planning}

\author{Jinhu Qi}
\email{jhqi25@cse.cuhk.edu.hk}
\affiliation{
  \institution{The Chinese University of Hong Kong}
  \city{Hong Kong}
  \country{China}
}

\author{Wentao Zhang}
\email{p2522808@mpu.edu.mo}
\affiliation{
\institution{Macao Polytechnic University}
\city{Macau}
\country{China}
}

\author{Siu Man Ng}
\email{smng@link.cuhk.edu.hk}
\affiliation{
  \institution{The Chinese University of Hong Kong}
  \city{Hong Kong}
  \country{China}
}

\author{Feiyang Xu}
\email{1155250524@link.cuhk.edu.hk}
\affiliation{
  \institution{The Chinese University of Hong Kong}
  \city{Hong Kong}
  \country{China}
}

\author{Yanyu Chen}
\email{chenyanyu.cse@link.cuhk.edu.hk}
\affiliation{
  \institution{The Chinese University of Hong Kong}
  \city{Hong Kong}
  \country{China}
}

\author{Yaoman Li}
\email{ymli@link.cuhk.edu.hk}
\affiliation{
  \institution{The Chinese University of Hong Kong}
  \city{Hong Kong}
  \country{China}
}

\author{Irwin King}
\email{king@cse.cuhk.edu.hk}
\affiliation{
  \institution{The Chinese University of Hong Kong}
  \city{Hong Kong}
  \country{China}
}

%\renewcommand{\shortauthors}{Qi et al.}

\begin{abstract}
Travel planning is a demanding stress test for tool-using LLM agents: a usable itinerary is a single artifact that must be right along many axes at once---every flight, hotel, and attraction must exist and be bookable, the days must be physically traversable, the total must clear a budget, and the plan must serve a traveler whose needs are only partly stated. Existing agent benchmarks reward these properties one at a time and grade the final output with soft or LLM-judged rubrics, which cannot \emph{certify} that a returned plan is executable and are neither reproducible nor auditable. We introduce \textbf{TREK} (Travel Reasoning and Evaluation Kit), a benchmark for \emph{feasible itinerary synthesis}: producing a \emph{single} plan that is jointly constraint-correct, hallucination-free, spatio-temporally executable, budget-valid, and responsive to the traveler's \emph{unstated} persona needs. TREK comprises 800 multi-constraint tasks---533 feasible and 267 provably infeasible with typed route/entity/budget causes---over a synthetic, internally consistent knowledge base of $212{,}530$ records across 375 cities and 13 personas, served through a production-style tool sandbox of validated RESTful APIs. Every task is scored by a fully deterministic, rule-based evaluator with no LLM judge and ships a human-verified gold reference that scores a perfect $1.0$ under that same evaluator, so the ceiling is demonstrably achievable and every remaining gap is an agent limitation rather than scorer strictness. Evaluating 15 LLM agents across nine constraint dimensions, we find that even the strongest (GPT-5.6) produces a fully-feasible plan on only \textbf{46.2\%} of solvable tasks, with a median of \textbf{6.6\%} and a floor of \textbf{0.0\%}; satisfying travelers' \emph{unstated} needs emerges as the universal bottleneck, unsolved even at the frontier. We release the dataset, tool sandbox, deterministic evaluator, and agent code as a fully reproducible benchmark.\footnote{\url{https://github.com/TonyQJH/TREK-A-Travel-Reasoning-and-Evaluation-Kit-for-LLM-Agents-in-Complex-Trip-Planning}}
\end{abstract}





\maketitle

% =============================================================================
% Main Content Sections
% =============================================================================
% =============================================================================
% Section 1: Introduction
% =============================================================================
\section{Introduction}

\begin{figure}[t]
  \centering
  \includegraphics[width=0.34\textwidth]{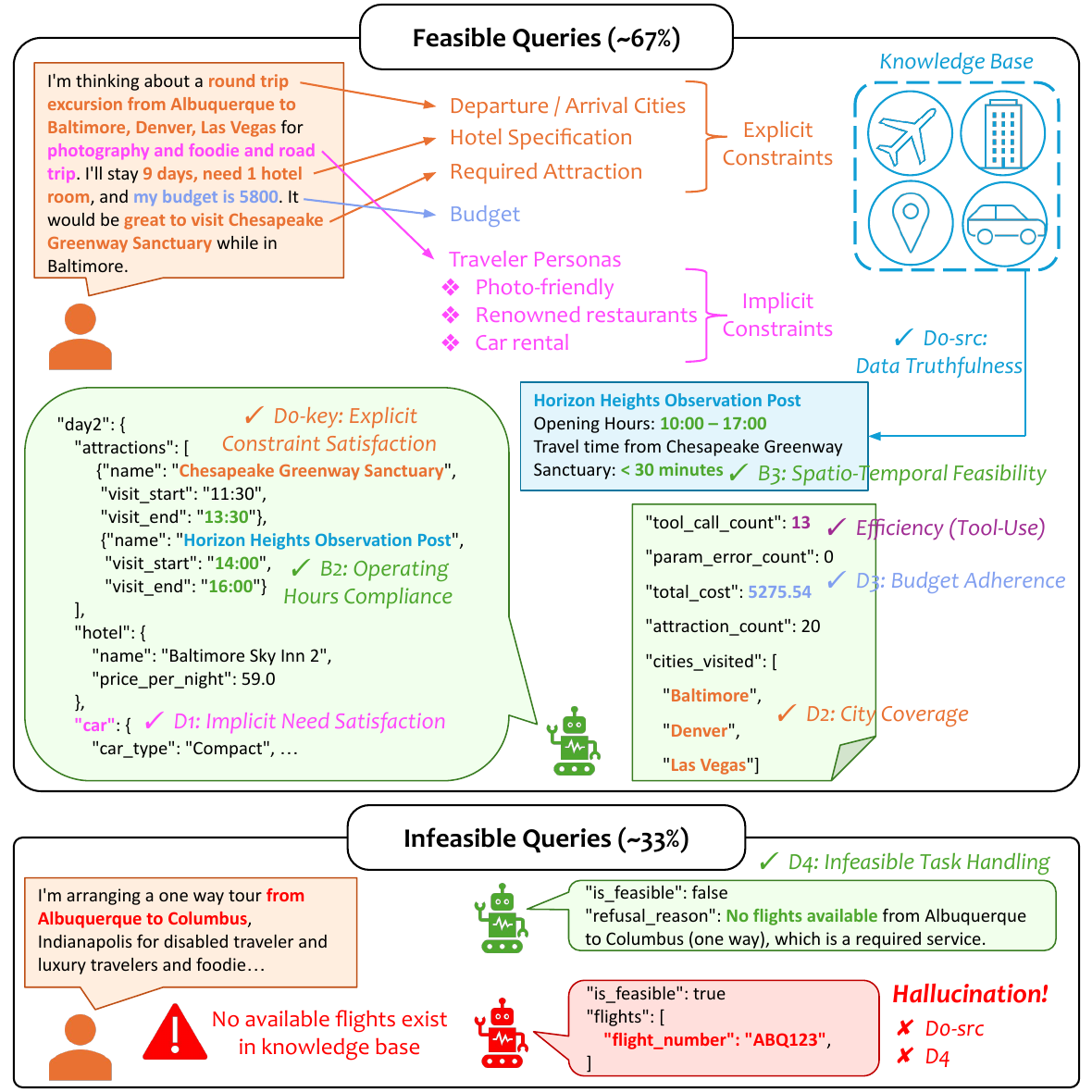}
  \caption{Examples of feasible and infeasible queries in TREK. Agents interact with a production-style tool sandbox to construct travel plans, which a fully deterministic, no-LLM-judge evaluator scores along nine constraint dimensions.}
  \Description{Of the 800 tasks in TREK, 533 are feasible and 267 are provably infeasible.}
  \label{fig:example}
\end{figure}

Large language model (LLM) agents are moving from chat transcripts into production, where they must not merely \emph{propose} an action but \emph{return an artifact that executes}~\cite{ReAct,Toolformer,toolllm}. Travel planning is the sharpest stress test for this shift. A deployable trip plan is a single object that must be right along many axes at once: every flight, hotel, and attraction must exist and be bookable, the days must be physically traversable, the total must clear a budget, and the whole must serve a traveler whose needs are only partly spoken. Getting any one axis right is routine for a modern tool-using agent; getting all of them right \emph{simultaneously}, in one plan, is the capability we isolate and measure---and one today's agents largely lack.

We name this capability \textbf{feasible itinerary synthesis}: emitting a \emph{single} plan that is jointly (a)~\textbf{constraint-correct} against the explicit request, (b)~\textbf{hallucination-free} (every entity resolves to a knowledge-base record), (c)~\textbf{spatio-temporally executable} (same-day travel is physically reachable in time), (d)~\textbf{budget-valid}, and (e)~\textbf{responsive to the traveler's \emph{unstated} persona needs}---all five holding \emph{simultaneously, in one artifact}.

\textbf{The measurement gap.} Prior agent and tool benchmarks reward these properties \emph{one at a time}, or one tool-call at a time~\cite{AgentBench,toolllm}, and grade the final output with soft rubrics or an LLM judge. First, no soft or LLM-judged score can \emph{certify} that a returned plan is executable end to end; such scores are neither reproducible nor auditable, and an agent can score well while returning a plan no traveler could follow. Second, satisfying five feasibility properties separately says nothing about satisfying them \emph{jointly in one artifact}---what deployment demands. Feasibility is a conjunction; measuring it demands a scorer that checks that conjunction deterministically. The bottleneck is therefore measurement itself: we cannot trust a number unless the scorer is exact and its ceiling reachable.

\textbf{TREK.} We introduce \textbf{TREK} (Travel Reasoning and Evaluation Kit), a benchmark built around \emph{trustworthy measurement} of joint feasibility. Agents act inside a production-\emph{style} tool sandbox --- RESTful APIs with strict parameter validation, semantic-search endpoints, and structured JSON errors --- over a knowledge base of \textbf{212{,}530} records spanning flights, hotels, attractions, and car rentals across \textbf{375} cities, instantiated over \textbf{13} traveler personas. The data is \emph{synthetic and structurally consistent by construction} --- not as a compromise, but because internal consistency is what lets us compute an exact ground truth and guarantee a reachable ceiling, guarantees no scraped, drifting corpus can offer. We do not claim the data is real-world-collected; we claim it is \emph{controlled}, the stronger property for certification.

Two trust properties form the paper's spine. \textbf{(1) A fully deterministic, rule-based evaluator with no LLM judge.} Scoring is bit-reproducible and free to re-run. Even implicit-need satisfaction is deterministic: the \emph{D1} scorer is a set-intersection of each persona's required facilities against the corresponding KB facility fields, not a learned or embedding-based match. (Agents may \emph{use} semantic search as a tool; the \emph{scoring} never does.) \textbf{(2) An achievable, human-verified gold.} Each of the \textbf{800} tasks ships with a gold reference---a feasible itinerary, or for an infeasible task the correctly-typed refusal---human-vetted by a 15-annotator panel, that scores \textbf{1.0} under that same evaluator. The ceiling is therefore \emph{demonstrably reachable}: any gap between an agent and 1.0 reflects the agent (model and fixed harness), not scorer strictness.

Determinism alone is not new; ChinaTravel~\cite{chinatravel} already scores against a deterministic DSL. TREK's novelty is the \emph{combination}: a deterministic no-judge evaluator, a human-verified gold that demonstrably achieves $1.0$, typed infeasibility as a first-class label, and a production-style tool sandbox, wired together to certify \emph{joint} feasibility against a demonstrably attainable ceiling. The \textbf{800} tasks split into \textbf{533} feasible and \textbf{267} \emph{provably} infeasible instances (Figure~\ref{fig:example}), the latter carrying \emph{typed} infeasibility labels (route, entity, budget) so that correctly \emph{refusing} an impossible request is a first-class, scored behavior.

\textbf{The gap.} Our headline metric, the \textbf{task-perfect rate}, is the fraction of tasks an agent solves on \emph{every} applicable one of \textbf{nine} constraint dimensions, reported feasible-only over the \textbf{533} solvable tasks. Across \textbf{15} LLM agents, the strongest (GPT-5.6) produces a fully-feasible plan on only \textbf{46.2\%} of solvable tasks; the \emph{median} of the 15 reaches just \textbf{6.6\%}, and the floor is \textbf{0.0\%}. TREK's top score is thus hard-but-in-band --- not a broken one (\S\ref{sec:related} calibrates against $\tau$-bench and related agentic benchmarks) --- with headroom to a \emph{demonstrably} reachable 1.0 the phenomenon we study.

\textbf{Research questions and findings.} We organize the study around three questions, resolving each where it is asked:
\begin{itemize}[leftmargin=*, itemsep=2pt]
  \item \textbf{RQ1: Can frontier LLM agents synthesize a fully-feasible itinerary?} \emph{No} --- 46.2\% at the top, 6.6\% at the median, 0.0\% at the floor.
  \item \textbf{RQ2: Where does feasible synthesis break down, and how does the bottleneck shift with capability?} Implicit-need satisfaction (D1) is the \emph{universal bottleneck}: it is the sole dimension the frontier model still fails at scale, clearing it on only \textbf{46.3\%} of applicable tasks while clearing every other dimension on at least \textbf{86.7\%} --- and a top-two failure for every one of the 15 agents. The bottleneck \emph{rises with capability}: weak models fail every dimension, while the frontier fails essentially only this one. Spatio-temporal reachability (B3, same-day travel-time feasibility) is the \emph{planner/non-planner watershed}, ranging from 13.3\% failure at the top to 93.8\% at the bottom. Only city-\emph{ordering} improves with capability; the task-level multi-city penalty persists, so multi-city planning is far from solved.
  \item \textbf{RQ3: Does more deliberation or compute buy more feasibility?} \emph{Not measurably.} In the one instruct/reasoning pair our set permits, the reasoning variant \emph{underperforms} --- a reasoning-vs-controllability tension under strict tool schemas (a cross-version observation, not a perfectly controlled pair). Accuracy also does not track spend: GPT-5.6 tops the board at the lowest per-query token cost among the top-scoring agents, while several weak models spend 5--7$\times$ the tokens for single-digit scores (tokens are provider-reported; wall-clock is a usage-cost proxy, not clean compute).
\end{itemize}

\noindent\textbf{Contributions.}
\begin{enumerate}[leftmargin=*, itemsep=2pt]
  \item A dataset of \textbf{800} joint-feasibility travel-planning tasks over a \textbf{212{,}530}-record synthetic knowledge base (\textbf{375} cities, \textbf{13} personas), served through a production-style tool sandbox of validated RESTful and semantic-search APIs with structured JSON errors.
  \item A fully deterministic, no-LLM-judge evaluator paired with a human-verified gold reference that demonstrably scores \textbf{1.0} on all 800 tasks, making the ceiling bit-reproducible and demonstrably achievable so that every gap is attributable to the agent.
  \item The \textbf{task-perfect} headline metric over nine constraint dimensions, plus typed (route / entity / budget) infeasibility as a first-class, scored evaluation dimension via the 267 provably-infeasible tasks.
  \item A \textbf{15-model} study establishing the top/median/floor gap and three findings: D1 as the capability-rising universal bottleneck, B3 as the planner watershed, and reasoning not helping in the one pair we can test while accuracy does not track cost.
\end{enumerate}

% =============================================================================
% Section 2: Related Work
% =============================================================================
\section{Related Work}
\label{sec:related}

\subsection{Travel-Planning Benchmarks}

Travel planning is a popular stress test for long-horizon, tool-using agents: a usable itinerary must jointly satisfy many heterogeneous constraints. TravelPlanner~\cite{travelplanner} is the most direct predecessor---an agent queries six travel databases and emits one multi-day plan under environmental, commonsense, and hard constraints---and its two-stage GPT-4-Turbo agent attains only $0.6\%$ final pass. But its free-form output is first \emph{structured by GPT-4-Turbo} before scripted checks, so the score is not parser-free, and it evaluates neither implicit preferences nor typed infeasibility. ChinaTravel~\cite{chinatravel} broadens the setting to authentic Chinese multi-day requests and validates plans with an \emph{executable compositional DSL} (a deterministic symbolic evaluator), its strongest agent neuro-symbolic. As our closest deterministic relative, it is what TREK builds \emph{beyond}: it supplies no gold \emph{proven} to attain the maximum, no \emph{typed} infeasibility, no production-style API sandbox, and no \emph{scored} efficiency axis. NATURAL PLAN~\cite{naturalplan} scores trip, meeting, and calendar planning by exact match but supplies tool outputs in-context rather than an interactive sandbox, and all models fall below $5\%$ on ten-city trip planning.

The now-crowded domain sharpens rather than dilutes our positioning; several recent efforts share our name or scope and must be distinguished. The concurrent ACL~2026 \emph{TravelBench}~\cite{travelbench_acl} (Cheng et al.) is closest in name and domain yet \emph{near-orthogonal}: it grades multi-turn \emph{dialogue} quality over real Amap logs with an \emph{LLM rubric and meta-judge}, marking infeasibility with descriptive interaction-boundary categories (missing-info/-tool/-intent). TREK instead certifies a \emph{single executable plan} deterministically---no LLM judge---against a gold that demonstrably scores $1.0$, with \emph{typed} route/entity/budget infeasibility and \emph{rule-based} implicit-need scoring. The KDD~2026 \emph{TravelEval}~\cite{traveleval} (Chen et al.) is a simulation-based whole-plan evaluator over multiple travel-quality dimensions, but its scores are neither bit-reproducible nor paired with a provably-maximal gold, and it neither \emph{types} infeasibility nor exposes a validated tool sandbox; one further unrelated work also uses the \emph{TravelBench} name~\cite{travelbench_lowresource}. A parallel wave studies complementary axes---spatio-temporal coherence and personalization~\cite{tripcraft,triptailor,tprag,agenttravel}, implicit or profile-conditioned preferences~\cite{retail_travel,personal_travel_solver,tripplus,grouptravelbench}, simulation-scored whole-plan dynamics~\cite{travelsim}, and disruption-driven replanning~\cite{triptide}. To our knowledge, no prior travel benchmark---including ChinaTravel, TravelEval, TravelBench, and this parallel wave---couples a fully deterministic, no-LLM-judge evaluator with a \emph{human-verified gold that demonstrably attains the maximum}, \emph{typed} route/entity/budget infeasibility, and an explicit efficiency axis over a single, jointly-feasible itinerary (Table~\ref{tab:benchmarks}).

% Table 1: TREK vs. related benchmarks. Facts sourced from each paper (deep-research survey, 2026-07-25).
\begin{table*}[t]
  \centering
  \footnotesize
  \setlength{\tabcolsep}{4pt}
  \caption{TREK versus representative planning and agent benchmarks.
  \textbf{Scoring}: rule-based (\emph{Rules}, no LLM judge), LLM-structuring-then-rules (\emph{Ext.+Rules}), executable \emph{DSL}, simulation (\emph{Sim.}), execution (\emph{Exec.}), exact \emph{Match}, or LLM \emph{Judge} (DSL/Sim.\ are also judge-free; TREK's differentiator is the \emph{combination}, not determinism alone).
  \textbf{API}: the agent acts through a typed tool sandbox with validated parameters and structured errors (vs.\ free-form function calls or browser actions). \textbf{Gold\,$=$\,1}: a reference solution demonstrably attains the evaluator maximum on every task ($-$: provided but not shown maximal; \xmark: none). \textbf{Impl.}: whether \emph{unstated} preferences are scored and \emph{how}---\emph{Det.} (deterministically, e.g., facility set-intersection), \emph{Judge} (LLM rubric), or $-$ (not separately specified). \textbf{Infeas.}: \emph{Typed} (route/entity/budget proof) vs.\ \emph{Untyped} vs.\ \xmark. \textbf{Effic.}: an explicit tool-use efficiency/cost metric is scored.
  To our knowledge, only TREK combines a single-plan output, purely rule-based scoring, a typed API sandbox, a verified achievable gold, implicit-need scoring, typed infeasibility, and efficiency scoring.}
  \label{tab:benchmarks}
  \begin{tabular*}{\textwidth}{@{\extracolsep{\fill}} llll cccl cl @{}}
    \toprule
    \textbf{Benchmark} & \textbf{Domain} & \textbf{Output} & \textbf{Scoring} & \textbf{API} & \textbf{Gold\,$=$\,1} & \textbf{Impl.} & \textbf{Infeas.} & \textbf{Effic.} & \textbf{Scale} \\
    \midrule
    \textbf{TREK (Ours)} & Travel & Plan & Rules & \cmark & \cmark & Det. & Typed & \cmark & 800 \\
    \midrule
    TravelPlanner~\cite{travelplanner}      & Travel          & Plan   & Ext.+Rules & \xmark & $-$    & \xmark & \xmark   & \xmark & 1{,}225 \\
    ChinaTravel~\cite{chinatravel}          & Travel          & Plan   & DSL        & \xmark & $-$    & Det.   & \xmark   & \xmark & 1{,}154 \\
    TravelEval~\cite{traveleval}            & Travel          & Plan   & Sim.       & \xmark & $-$    & $-$    & \xmark   & \xmark & 1{,}150 \\
    TravelBench~\cite{travelbench_acl}      & Travel          & Turn   & Judge      & \cmark & \xmark & Judge  & Untyped  & \xmark & 1{,}100 \\
    NATURAL PLAN~\cite{naturalplan}         & Trip/Mtg/Cal.   & Plan   & Match      & \xmark & \cmark & \xmark & \xmark   & \xmark & 3{,}600 \\
    $\tau$-bench~\cite{taubench}            & Airline/Retail  & Turn   & Exec.      & \cmark & \xmark & \xmark & \xmark   & \xmark & 165 \\
    WebArena~\cite{zhou2023webarena}        & Web             & Step   & Exec.      & \xmark & \xmark & \xmark & Untyped  & \xmark & 812 \\
    GAIA~\cite{gaia}                        & General         & Answer & Match      & \xmark & $-$    & \xmark & \xmark   & \xmark & 466 \\
    ToolBench~\cite{toolllm}                & Tool API        & Step   & Judge      & \cmark & \xmark & \xmark & Untyped  & \xmark & 16{,}464 APIs \\
    \bottomrule
  \end{tabular*}
\end{table*}

\subsection{Realistic Agent Environments and Verifiable Planning}

Beyond travel, benchmarks establish the value of stateful environments and execution-based scoring. $\tau$-bench~\cite{taubench} evaluates tool-agent-user interaction against a deterministic database end-state; WebArena~\cite{zhou2023webarena} and OSWorld~\cite{osworld} score real web and desktop workflows by execution; GAIA~\cite{gaia} checks normalized answers to cross-tool research questions; and AgentBench~\cite{AgentBench}, ToolBench~\cite{toolllm}, API-Bank~\cite{API-Bank}, and SWE-bench~\cite{swebench} probe tool orchestration and code-issue resolution. Best-system success spans a broad band---for calibration, archived $\tau$-bench reports $46.0\%$ airline / $69.2\%$ retail pass---so TREK's $46.2\%$ top full-plan-perfect rate on feasible tasks (RQ1, \S\ref{ssec:rq1}) places it among difficult-but-tractable agentic tasks rather than broken ones. PlanBench~\cite{planbench} and the ``LLMs cannot plan'' line~\cite{planningabilities,llmmodulo} argue that linguistic plausibility is a weak proxy for formal executability and advocate external validators or solver-centered architectures; formal-verification and planner-augmented travel systems~\cite{formaltravel,trippal,robustllmmodulo,ttg} confirm that solver backends lift TravelPlanner scores well above its neural baseline. TREK inherits this execution-first philosophy but differs in \emph{what} it certifies: where these score a database end-state, a web step, or a normalized answer, TREK certifies one \emph{multi-constraint artifact}'s \emph{joint} feasibility against a demonstrably-reachable ceiling, with facts only from validated APIs (\S\ref{ssec:api-env}).

\subsection{Deterministic Evaluation and Implicit Needs}

Our use of \emph{no} LLM judge is grounded in growing evidence that LLM judges are biased and unstable: they exhibit position, verbosity, self-preference, and authority biases and drift across prompts and runs~\cite{mtbench,notfairevaluators,stureborgjudge,positionbias,selfpreference,justiceprejudice,judgebench}, motivating program-based evaluation for objectively decidable properties~\cite{pajama}. Because every TREK dimension is objectively checkable against the knowledge base, we score with deterministic rules and reserve human judgment for the gold (\S\ref{sec:evaluation}). Determinism alone buys only a \emph{reproducible} ceiling; TREK pairs it with a gold \emph{proven} to hit the maximum on all nine dimensions at once, so any agent-to-ceiling gap is attributable to the agent, not the scorer. Orthogonally, TREK evaluates \emph{unstated} persona needs, connecting to work on personalization and preference following~\cite{lamp,prefeval,aloe} and accessibility-aware assistance~\cite{accesseval}---a capability explicit-constraint benchmarks rarely isolate, and which we find the single hardest dimension even for frontier agents (the universal bottleneck of \S\ref{ssec:rq2}).

\subsection{Reasoning vs.\ Instruct Models}

TREK's production-style sandbox lets us ask a question the implicit-need bottleneck sharpens---does more deliberation help?---where the two dominant model recipes pull in opposite directions. Instruction tuning and alignment improve output controllability and format adherence~\cite{ouyang2022instructgpt}, whereas chain-of-thought and reasoning-specialized models invest extra computation in deliberation for complex multi-step problems~\cite{CoT,openai2024o1systemcard,guo2025deepseek}. Under strict tool schemas, longer deliberation can induce format drift, redundant calls, and ``overthinking''~\cite{internalbiasreasoningmodels}. Prior benchmarks rarely contrast reasoning and instruct behavior under identical production-style interfaces; our experiments (\S\ref{ssec:rq3}) surface a reasoning-vs-controllability tension in this regime---with higher token spend not tracking higher task accuracy---though as a single-pair, cross-version observation rather than a perfectly controlled ablation.

% =============================================================================
% Section 3: The TREK Benchmark
% =============================================================================
\section{The TREK Benchmark}
\label{sec:dataset}

\begin{figure*}[t]
  \centering
  \includegraphics[width=\textwidth]{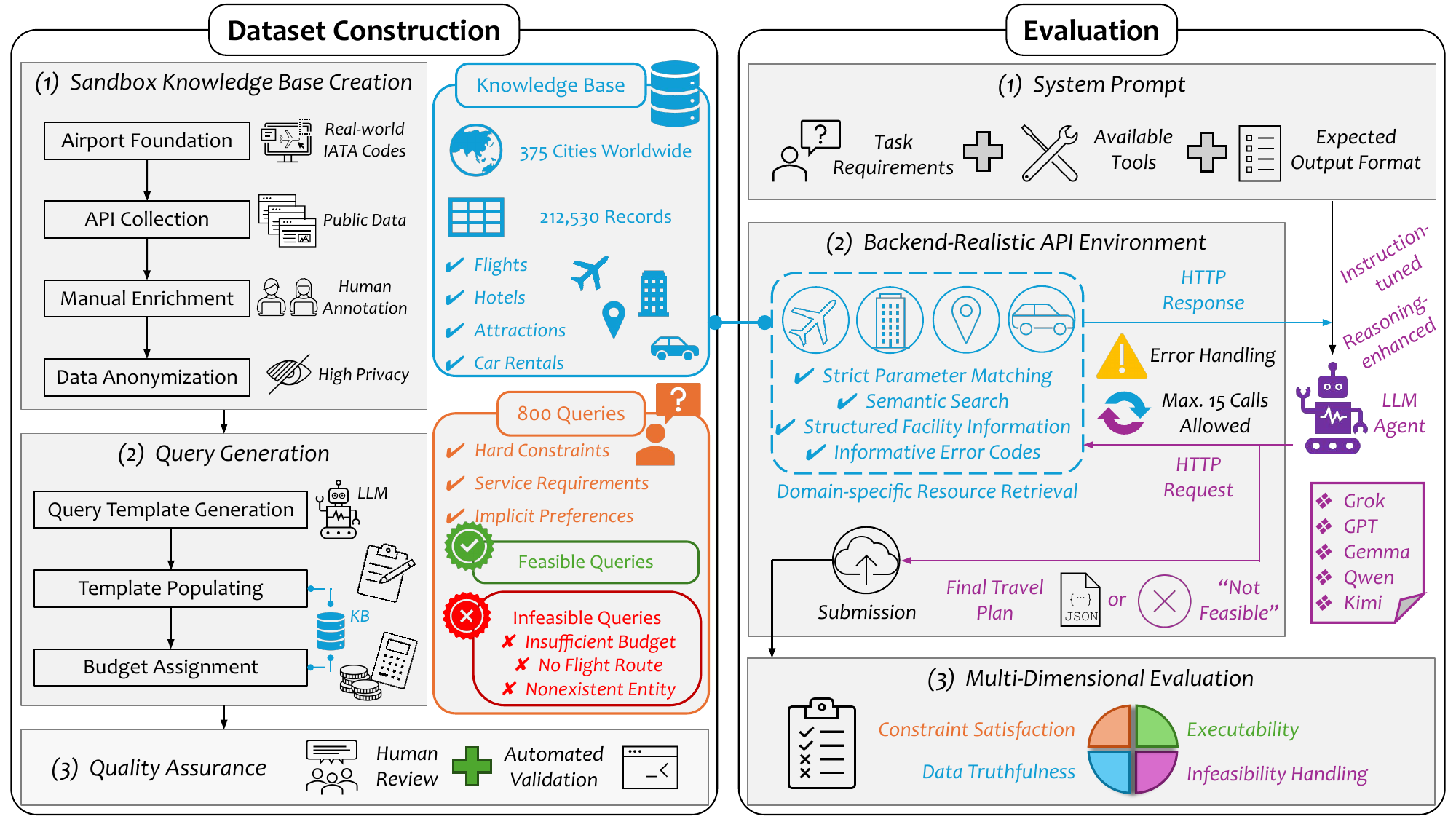}
  \caption{Overview of TREK. A deterministic pipeline (left) builds a synthetic, internally consistent knowledge base, populates pure templates into 800 tasks whose feasibility labels are correct \emph{by construction}, and ships a human-verified gold itinerary for every task. Agents (right) interact with a production-style tool sandbox and are graded by a fully deterministic, multi-dimensional evaluator with no LLM judge.}
  \Description{A deterministic three-stage pipeline builds a synthetic knowledge base, generates 800 tasks with construction-correct labels, and verifies a gold plan for each; agents then interact with a tool sandbox and are scored deterministically across nine dimensions.}
  \label{fig:main}
\end{figure*}

TREK comprises \textbf{800} multi-constraint travel-planning tasks---\textbf{533} feasible and \textbf{267} provably infeasible---grounded in a synthetic, internally consistent knowledge base of \textbf{212{,}530} records (flights, hotels, attractions, car rentals) over \textbf{375} cities, and served through a production-style tool sandbox (Figure~\ref{fig:main}). We motivate the sandbox knowledge base (\S\ref{ssec:kb}), then formalize the task (\S\ref{ssec:task-def}), query generation (\S\ref{ssec:query-gen}), and typed infeasible tasks (\S\ref{ssec:infeasible}); we close with human verification (\S\ref{ssec:qa}) and the API environment (\S\ref{ssec:api-env}).

% -----------------------------------------------------------------------------
% 3.1 Sandbox Knowledge Base
% -----------------------------------------------------------------------------
\subsection{A Synthetic, Internally Consistent Sandbox}
\label{ssec:kb}

TREK is grounded in a \emph{synthetic} knowledge base, generated by a deterministic build script that reads only fixed seed inputs---a deliberate design choice, not a compromise. Certifying that an itinerary is executable requires stable ground truth: prices, coordinates, room types, facility lists, and flight routes must be mutually coherent so a rule-based scorer can decide feasibility and a single reference itinerary can attain the maximum score. A scraped, drifting corpus offers no such guarantee, and redistributing real listings raises copyright and privacy concerns. We \emph{do not} claim the data is real; we claim it is \emph{controlled}---the stronger property for reproducible certification. For geographic realism, the city-to-airport scaffold uses the public OurAirports reference~\cite{ourairports}; all bookable content (schedules, prices, amenities, ratings) is synthesized and internally reconciled.

\noindent The knowledge base holds $212{,}530$ mutually-consistent synthetic records---$107{,}195$ flights, $39{,}396$ hotels, $55{,}814$ attractions, and $10{,}125$ car rentals---with $392$ airports\footnote{Counted as the distinct airports appearing in the flight network; a city may be served by more than one.} serving the $375$ cities (per-domain breakdown in Appendix Table~\ref{tab:kb-stats}).

\paragraph{Schema conventions.}
Each domain follows a fixed schema (full field list in Appendix~\ref{app:facilities}). Two conventions matter for scoring: quality \emph{ratings} lie on a $[0,10]$ scale, whereas hotel \emph{star} classes lie on $\{1,\dots,5\}$; and a hotel record stores a single nightly price for a room seating two guests, from which multi-room, multi-night costs are computed deterministically. These conventions are intentional and applied identically by the generator, sandbox, and scorer.

% -----------------------------------------------------------------------------
% 3.2 Task Definition
% -----------------------------------------------------------------------------
\subsection{Task Formulation}
\label{ssec:task-def}

We cast travel planning as a constraint-satisfaction problem: given a natural-language query $q$, an agent must return either a structured itinerary $P$ satisfying all constraints, or a refusal correctly diagnosing why the task is infeasible. TREK is a collection of query--annotation pairs:
\begin{equation}
    \mathcal{D} = \{(q_1, A_1), \ldots, (q_N, A_N)\}, \quad N = 800 .
    \label{eq:dataset-def}
\end{equation}
Each annotation decomposes into hard constraints, service requirements, and implicit persona keywords:
\begin{equation}
    A_i = \Bigg( \underbrace{\underbrace{\{pn, bg, dy\}^{(i)}}_{C_{\text{hard}}^{(i)}} , \underbrace{\{f, h, c, a\}^{(i)}}_{\textit{Req}^{(i)}}}_{C_{\text{explicit}}^{(i)}} , \underbrace{[kw_1, kw_2, \ldots]^{(i)}}_{K_{\text{implicit}}^{(i)}} \Bigg),
    \label{eq:annotation-def}
\end{equation}
where $C_{\text{hard}}=\{pn,bg,dy\}$ is person count, budget, and trip days; $\textit{Req}=\{f,h,c,a\}$ are the required services (flights, hotels, cars, attractions); and $K_{\text{implicit}}$ holds persona keywords encoding latent needs. Every task requires all four services; symbol definitions are in Appendix~\ref{app:symbol-def}.

\paragraph{Personas and implicit needs.}
We define \textbf{13} traveler personas (e.g., ``with children'', ``elderly travelers'', ``disabled traveler'', ``luxury travelers''), each mapped deterministically to required facilities (complete table in Appendix~\ref{app:implicit-requirements}). A task carries $0$--$3$ personas (mean $1.55$; $120$ tasks carry none), so implicit-need satisfaction is scored only where a persona applies. We exclude only \emph{logically impossible} persona combinations---party-size contradictions from the stated person count (e.g., ``solo women'' implies exactly one traveler) and the single same-axis conflict (``fast-paced budget travel'' vs.\ ``luxury travelers''). An earlier hand-written rule set of nineteen ``incompatible'' pairs was retired: an audit found both members of every such pair co-occurring on thousands of resources---as commonly as the pairs the same rules permitted---and several pairs encoded age- or disability-based stereotypes. Keeping only logically impossible exclusions removes that bias while preserving construction-time validity.

% -----------------------------------------------------------------------------
% 3.3 Query Generation
% -----------------------------------------------------------------------------
\subsection{Query Generation with Construction-Correct Labels}
\label{ssec:query-gen}

We use a \textbf{template-then-populate} approach decoupling surface diversity from factual grounding. A curated pool of \emph{pure} natural-language templates (containing no ordering cues) is populated by a deterministic, seeded script that samples valid entities and constraints from the knowledge base. Because population reads only the sandbox, every named entity, route, and price in a query resolves against the KB by construction.

\paragraph{Labels correct by construction.}
Feasibility labels are not annotated after the fact---they are \emph{guaranteed} at generation time. Each task's budget is derived from a reference plan priced by the \emph{same} cost model the scorer bills with: feasible tasks receive a budget above the achievable cost (tight or loose band), and budget-infeasible tasks a budget below the achievable floor. Every row is then re-scored by the authoritative evaluator before shipping, and only rows whose realized label matches their intended label are kept. Generation is fully deterministic (seeded RNG, reading only the versioned KB), so the task set is reproducible from source. Design details are in Appendix~\ref{app:query-generation}.

% -----------------------------------------------------------------------------
% 3.4 Typed Infeasible Tasks
% -----------------------------------------------------------------------------
\subsection{Typed Infeasible Tasks}
\label{ssec:infeasible}

To test whether agents refuse impossible requests rather than fabricate plans, \textbf{267} of the 800 tasks ($33.4\%$) are provably infeasible, balanced across three \emph{typed}, machine-checkable causes (\textbf{89} each):
\begin{itemize}[leftmargin=*, itemsep=1pt]
  \item \textbf{Entity} --- the query mandates a specifically named hotel, attraction, or car type that does not exist in any required city.
  \item \textbf{Route} --- a required direct leg between two cities is not served by any flight in the KB.
  \item \textbf{Budget} --- the budget is set strictly below the achievable cost floor for the trip.
\end{itemize}
A correct response must not merely refuse but \emph{diagnose the cause}. The evaluator re-derives the true cause independently from the task and the KB---checking a missing named entity, then a missing required leg, then the budget floor---rather than trusting a stored label, so ``I refuse'' scores strictly below a refusal that names the right reason (\S\ref{sec:evaluation}). Unlike broad capability-boundary categories, each cause is a verifiable proof about the world.

% -----------------------------------------------------------------------------
% 3.5 Quality Assurance
% -----------------------------------------------------------------------------
\subsection{Human Verification and Gold Itineraries}
\label{ssec:qa}

Every feasible task ships with a \emph{gold reference itinerary} that scores a perfect $1.0$ under the deterministic evaluator, and every infeasible task ships a gold \emph{typed refusal} that likewise scores $1.0$ by naming the correct cause---so the ceiling is provably achievable on all 800 tasks. These gold references were validated by a panel of \textbf{15 annotators} (10 PhD researchers, 5 travel-industry practitioners), who rated every feasible gold itinerary for realism and executability on a $1$--$5$ scale (mean $4.25/5$, $\approx$85\%); low-scored or disputed plans were revised on their feedback before release, so the gold is human-vetted, not a score-maximizing artifact. The panel also checked template diversity, entity-to-KB mappings, and persona-consistency, and automated scripts verified label correctness (re-scoring), KB back-tracing, and format consistency (protocol in Appendix~\ref{app:gold-protocol}).

% -----------------------------------------------------------------------------
% 3.6 API Environment
% -----------------------------------------------------------------------------
\subsection{Production-Style Tool Sandbox}
\label{ssec:api-env}

Unlike benchmarks built on simplified database lookups, TREK exposes production-style RESTful APIs that mirror real travel-service interfaces. We implement five endpoints: four for querying domain resources (flights, hotels, attractions, car rentals) and one for submitting the final plan. Each enforces strict parameter validation with required fields (origin/destination, dates, passenger counts), returns consistent JSON schemas with pagination, and emits structured errors the agent must interpret and act on. Full specifications are in Appendix~\ref{app:api-specs}.

\paragraph{Semantic search for implicit needs.}
Each query endpoint also returns structured facility information (e.g., amenity and service lists) and supports semantic search, which agents may use to \emph{find} resources matching a persona. Whether those needs are \emph{satisfied} is then scored deterministically by exact facility set-intersection---never by an LLM judge or embedding threshold---as detailed in Section~\ref{sec:evaluation}.

% =============================================================================
% Section 4: Evaluation Framework
% =============================================================================
\section{Evaluation Framework}
\label{sec:evaluation}

TREK is scored by a \emph{fully deterministic, rule-based evaluator with no LLM judge}: each dimension is an exact computation against the versioned knowledge base, making a submission's score bit-reproducible and free to re-run. We cover design principles (\S\ref{ssec:eval-principles}), the nine dimensions (\S\ref{ssec:scoring-dimensions}), hard gates (\S\ref{ssec:hard-gates}), and aggregation into the headline \emph{task-perfect rate} and a separate efficiency axis (\S\ref{ssec:composite-scores}).

\subsection{Design Principles}
\label{ssec:eval-principles}

\paragraph{No LLM judge.}
LLM judges are biased and unstable~\cite{mtbench,notfairevaluators,positionbias,judgebench} and cannot \emph{certify} that a plan is executable. Because every TREK property is objectively decidable against the KB, we score with deterministic rules and reserve human judgment for gold validation. The evaluator uses no embedding model, no learned threshold, and no network access.

\paragraph{Applicability is a property of the task, not the submission.}
A dimension is scored on a task iff the \emph{task} brings it into scope---e.g., implicit-need satisfaction only when the query carries a persona, multi-city coverage only when it names several cities. Applicability therefore never depends on what the agent emitted, so per-dimension denominators are a benchmark property, \emph{identical for every model}; each dimension's task count $n_{\text{scored}}$ is released with the evaluator output, so a reader can confirm models are averaged over the same tasks.

\paragraph{Achievable ceiling.}
Every task's gold reference scores $1.0$ on all four correctness categories under this evaluator (\S\ref{ssec:qa}), so the ceiling is provably attainable and every correctness gap is an agent limitation, not scorer strictness. Efficiency is a separate cost axis with no gold reference---a gold plan carries no tool-call trace---so it is excluded from this guarantee.

\subsection{Correctness Dimensions}
\label{ssec:scoring-dimensions}

We verify nine correctness dimensions in four categories (D2 contributes two, via its single- and multi-city coverage variants). Codes match the released evaluator's field names; each dimension returns a score in $[0,1]$ or \texttt{None} when inapplicable.

\paragraph{Constraint Satisfaction \{D0-key, D1, D2, D3\}.}
Two members are pass/fail. \emph{D0-key} requires \emph{every} explicitly requested element---named hotels, car types, attractions, flight endpoints, daily cars---plus the implied itinerary-completeness (all days spanned, lodging on every away-night, an activity in every stay city); \emph{D2} requires \emph{all} destination cities to be booked (reported as D2-single / D2-multi). The two graded members are \emph{D3} (budget: an exponential overspend penalty, $\beta{=}4$, so a $25\%$/$50\%$ overspend scores $0.37$/$0.14$) and \emph{D1} (implicit persona needs, by \textbf{deterministic facility set-intersection}---no embedding, no \emph{learned} threshold: each (persona, resource-type, city) cell scores $1$ iff the best booked resource carries a required facility, and D1 is the mean over cells; the \emph{luxury} and \emph{foodie} personas add a fixed star/rating quality test).

\paragraph{Truthfulness \{D0-src\}.}
\emph{D0-src} is \textbf{zero-tolerance and binary}: $1.0$ iff \emph{every} named entity resolves to a real KB record---a flight matched on number, endpoints, \emph{and} a consistent schedule; name-with-city for hotels/attractions; type-with-city for cars---else $0.0$. One fabrication zeroes the task, matching deployment, where an invented hotel breaks the trip.

\paragraph{Executability \{B2, B3\}.}
\emph{B2} is the fraction of attraction visits within opening hours (a compliant visit exists in the KB, fits inside its window, and spans its minimum duration). \emph{B3} is spatio-temporal feasibility: a same-day transition violates if its time gap is below the minimum door-to-door travel time from one canonical model---the faster of surface and air travel---\emph{shared verbatim} with the agent's \texttt{compute\_travel\_time} tool, so no agent is graded on a rule it could not query.

\paragraph{Infeasibility Handling \{D4\}.}
On the $267$ infeasible tasks, \emph{D4} is \textbf{all-or-nothing}: $1.0$ only for a refusal naming the \emph{correct typed cause} (entity / route / budget), which the evaluator re-derives independently, else $0.0$. Scored over the infeasible tasks only, it is a genuine headline rather than a near-constant $1.0$ padded by the feasible ones. Full formulas and denominators are in Appendix~\ref{app:scoring-details}.

\subsection{Hard Gates}
\label{ssec:hard-gates}

Some failures make an itinerary unusable or dishonest and carry no partial value; these are \emph{binary} gates, not graded dimensions. A submission fails validity if it (i)~books nothing, (ii)~invents any entity (a flight, hotel, attraction, or car absent from the KB, including a city-less entity), (iii)~schedules a \emph{majority} of its checkable visits outside opening hours, or (iv)~omits an entity the query explicitly mandated. The finer graded dimensions (D1, D3, B2, B3 in $[0,1]$) still feed the per-dimension \emph{failure decomposition} (Appendix~\ref{app:failure-rates}), which localizes \emph{where} a plan falls short even when a category records only pass/fail. These gates fire on $0$ of the $800$ gold plans.

\subsection{Aggregation and the Headline Metric}
\label{ssec:composite-scores}

\paragraph{Categories are all-or-nothing.}
Each of the four correctness categories is scored \textbf{all-or-nothing}: on a task the category is met only if \emph{every} applicable dimension in it scores $1.0$---so a plan that satisfies three of the four Satisfaction requirements still fails Satisfaction---and the category score is the fraction of applicable tasks that pass. This prevents one strong dimension from propping up a category, matching how a traveler judges a plan: a requirement group is either satisfied or it is not. (Truthfulness and Infeasibility each rest on a single all-or-nothing dimension; Infeasibility thus equals the infeasible task-perfect rate.) Because a deployable plan must be \emph{simultaneously} constraint-satisfying, truthful, executable, and correctly-refusing-when-impossible---conjunctive requirements, not tradeable features---we aggregate the four with a \textbf{geometric mean}:
\begin{equation}
    \text{Overall} = \Big(\textstyle\prod_{c \in \{\text{Sat, Truth, Exec, Infeas}\}} \bar{C}_c\Big)^{1/4}.
    \label{eq:geomean}
\end{equation}
An arithmetic mean would let perfect truthfulness compensate for an itinerary nobody can physically follow; the geometric mean lets no category hide behind the others (each category is floored at a small $\epsilon$ before the product, so a single zero keeps the score defined rather than annihilating discrimination). We do \emph{not} renormalize over available categories---an unmeasured category withholds the headline rather than inflating it.

\paragraph{Efficiency is a separate cost axis.}
Tool-call and token cost is reported as its own \emph{Efficiency} score, \emph{not} folded into the conjunctive headline: a correct itinerary produced expensively is still correct, so folding cost into correctness would wrongly make cheapness a precondition of being right. Efficiency compares the agent's tool-call count to a task-derived oracle minimum (penalizing both over- and under-calling, since under-calling means the plan was not grounded) and applies an exponential penalty for token overrun against a task-derived budget.

\paragraph{Headline: task-perfect rate.}
Our primary metric is binary per task and continuous across the benchmark, in the spirit of HumanEval pass@1: a task is \emph{perfect} iff it scores $1.0$ on \emph{every} applicable correctness dimension \emph{and} passes all hard gates. We report the \textbf{task-perfect rate} separately over the $533$ feasible tasks (the planning headline) and the $267$ infeasible tasks (the refusal headline), since blending executable-plan construction with near-saturated refusal detection flatters the number. The bar's position is not a cliff: requiring $\geq 0.95$ instead of $=1.0$ on every dimension moves at most one task for any model ($13$ of the $15$ are unchanged, the other $2$ shift by a single task, $\leq 0.2$\,pp). As a companion, the \emph{explicit-full rate} is the fraction of tasks on which the agent met \emph{every} required plan element ($\text{D0-key}=1.0$)---each explicit constraint together with the itinerary-completeness checks above---treating them as all-or-nothing.

% =============================================================================
% Section 5: Experiments
% =============================================================================
\section{Experiments}
\label{sec:experiments}

We evaluate 15 LLM agents on all 800 TREK tasks around three questions: whether current agents can synthesize a fully-feasible itinerary (RQ1, \S\ref{ssec:rq1}), where synthesis breaks down and how the bottleneck moves with capability (RQ2, \S\ref{ssec:rq2}), and whether more deliberation or compute buys more feasibility (RQ3, \S\ref{ssec:rq3}).

\subsection{Experimental Setup}
\label{ssec:exp-setup}

We evaluate \textbf{15} agents from ten vendors served through Amazon Bedrock (full list, request dialects, and token limits in Appendix Table~\ref{tab:model-specs}); GPT-5.6 is the one frontier closed model, and the Claude family is geo-blocked from our run location---an availability constraint, not a design choice. All run under one Bedrock-native function-calling agent (four search tools plus \texttt{submit\_plan}, capped at $15$ billable calls at temperature $0$; a free \texttt{compute\_travel\_time} helper exposes the \emph{exact} B3 travel-time model of \S\ref{ssec:scoring-dimensions}, so no agent is graded on a rule it could not query, and a lossless notebook retains every scorer-relevant field). Every task is graded by the deterministic evaluator of \S\ref{sec:evaluation}, under which the gold scores $1.0$.

% Main results — ALL-OR-NOTHING categories (each = fraction of tasks satisfying EVERY dimension in the group). Generated from trek_scores_binary/summary.json. Bold=best per column. Do not hand-edit numbers.
\begin{table*}[t]
  \centering \footnotesize \setlength{\tabcolsep}{3pt}
  \definecolor{gradgreen}{RGB}{56,142,60}
  \caption{Main results on TREK (15 agents, 800 tasks). \textbf{TP-feas}/\textbf{TP-inf} are the \emph{task-perfect} rate (all applicable dimensions $=1.0$ and all hard gates passed) over the 533 feasible and 267 infeasible tasks; TP-feas is the headline. \textbf{Sat/Tru/Exe/Inf} are the four correctness categories, each \emph{all-or-nothing}: a task counts only if it satisfies \emph{every} dimension in the group (Satisfaction: all explicit elements, implicit needs, cities, and budget; Executability: every visit in hours and every hop reachable). Since infeasibility handling is a single all-or-nothing check, \textbf{Inf} coincides with TP-inf. \textbf{Eff} is the standalone efficiency axis; \textbf{tok/q} is mean provider tokens per task. Rows sorted by TP-feas; \textbf{bold}=best per column; cell shading encodes value (darker~$=$~better). Gold scores $100$ on every correctness column (Eff and tok/q have no gold reference).}
  \label{tab:main_results}
  \begin{tabular*}{\textwidth}{@{\extracolsep{\fill}} l cc cccc c r @{}}
    \toprule
    & \multicolumn{2}{c}{\textbf{Task-Perfect}} & \multicolumn{4}{c}{\textbf{All-or-nothing categories}} & & \\
    \cmidrule(lr){2-3}\cmidrule(lr){4-7}
    \textbf{Model} & TP-feas & TP-inf & Sat & Tru & Exe & Inf & Eff & tok/q \\
    \midrule
    GPT-5.6 & \cellcolor{gradgreen!19}\textbf{46.2} & \cellcolor{gradgreen!41}\textbf{97.0} & \cellcolor{gradgreen!21}\textbf{50.7} & \cellcolor{gradgreen!40}\textbf{94.9} & \cellcolor{gradgreen!36}\textbf{86.3} & \cellcolor{gradgreen!41}\textbf{97.0} & \cellcolor{gradgreen!38}\textbf{90.4} & 46k \\
    Kimi-K2.5 & \cellcolor{gradgreen!7}17.1 & \cellcolor{gradgreen!39}93.6 & \cellcolor{gradgreen!14}32.8 & \cellcolor{gradgreen!39}92.9 & \cellcolor{gradgreen!19}44.8 & \cellcolor{gradgreen!39}93.6 & \cellcolor{gradgreen!33}78.0 & 92k \\
    GLM-5 & \cellcolor{gradgreen!6}13.1 & \cellcolor{gradgreen!38}91.0 & \cellcolor{gradgreen!13}31.3 & \cellcolor{gradgreen!38}90.1 & \cellcolor{gradgreen!17}41.5 & \cellcolor{gradgreen!38}91.0 & \cellcolor{gradgreen!35}82.4 & 78k \\
    Gemma-4-31B & \cellcolor{gradgreen!5}12.0 & \cellcolor{gradgreen!35}84.3 & \cellcolor{gradgreen!11}25.5 & \cellcolor{gradgreen!36}85.7 & \cellcolor{gradgreen!12}28.7 & \cellcolor{gradgreen!35}84.3 & \cellcolor{gradgreen!14}32.9 & 203k \\
    GLM-4.7 & \cellcolor{gradgreen!5}11.8 & \cellcolor{gradgreen!36}86.5 & \cellcolor{gradgreen!13}29.8 & \cellcolor{gradgreen!37}87.1 & \cellcolor{gradgreen!14}34.3 & \cellcolor{gradgreen!36}86.5 & \cellcolor{gradgreen!29}70.1 & 105k \\
    GPT-OSS-120B & \cellcolor{gradgreen!4}10.5 & \cellcolor{gradgreen!32}75.7 & \cellcolor{gradgreen!8}19.9 & \cellcolor{gradgreen!34}81.6 & \cellcolor{gradgreen!17}40.2 & \cellcolor{gradgreen!32}75.7 & \cellcolor{gradgreen!27}64.4 & 112k \\
    Grok-4.3 & \cellcolor{gradgreen!3}7.5 & \cellcolor{gradgreen!36}86.9 & \cellcolor{gradgreen!10}23.1 & \cellcolor{gradgreen!32}76.2 & \cellcolor{gradgreen!7}15.8 & \cellcolor{gradgreen!36}86.9 & \cellcolor{gradgreen!30}71.0 & \textbf{32k} \\
    Kimi-K2-Thinking & \cellcolor{gradgreen!3}6.6 & \cellcolor{gradgreen!37}87.3 & \cellcolor{gradgreen!10}22.7 & \cellcolor{gradgreen!36}86.9 & \cellcolor{gradgreen!11}26.5 & \cellcolor{gradgreen!37}87.3 & \cellcolor{gradgreen!32}76.2 & 68k \\
    Mistral-Large-3 & \cellcolor{gradgreen!2}5.8 & \cellcolor{gradgreen!31}73.8 & \cellcolor{gradgreen!7}15.8 & \cellcolor{gradgreen!34}81.2 & \cellcolor{gradgreen!12}29.6 & \cellcolor{gradgreen!31}73.8 & \cellcolor{gradgreen!30}71.6 & 56k \\
    DeepSeek-V3.2 & \cellcolor{gradgreen!2}4.9 & \cellcolor{gradgreen!38}91.0 & \cellcolor{gradgreen!13}30.2 & \cellcolor{gradgreen!40}94.6 & \cellcolor{gradgreen!6}14.1 & \cellcolor{gradgreen!38}91.0 & \cellcolor{gradgreen!12}27.8 & 270k \\
    GPT-OSS-20B & \cellcolor{gradgreen!2}3.9 & \cellcolor{gradgreen!31}74.5 & \cellcolor{gradgreen!5}12.0 & \cellcolor{gradgreen!28}65.5 & \cellcolor{gradgreen!11}26.1 & \cellcolor{gradgreen!31}74.5 & \cellcolor{gradgreen!16}38.0 & 186k \\
    Nova-2-Lite & \cellcolor{gradgreen!1}2.6 & \cellcolor{gradgreen!12}28.8 & \cellcolor{gradgreen!4}8.4 & \cellcolor{gradgreen!33}77.5 & \cellcolor{gradgreen!8}18.2 & \cellcolor{gradgreen!12}28.8 & \cellcolor{gradgreen!11}26.5 & 324k \\
    Llama-4-Maverick & \cellcolor{gradgreen!0}0.4 & \cellcolor{gradgreen!26}61.8 & \cellcolor{gradgreen!1}3.4 & \cellcolor{gradgreen!27}65.3 & \cellcolor{gradgreen!10}22.9 & \cellcolor{gradgreen!26}61.8 & \cellcolor{gradgreen!20}47.1 & 85k \\
    Qwen3-Next-80B & \cellcolor{gradgreen!0}0.4 & \cellcolor{gradgreen!10}23.2 & \cellcolor{gradgreen!3}6.9 & \cellcolor{gradgreen!25}60.0 & \cellcolor{gradgreen!2}3.9 & \cellcolor{gradgreen!10}23.2 & \cellcolor{gradgreen!15}34.8 & 229k \\
    Nova-Pro & \cellcolor{gradgreen!0}0.0 & \cellcolor{gradgreen!25}59.9 & \cellcolor{gradgreen!2}5.1 & \cellcolor{gradgreen!30}72.2 & \cellcolor{gradgreen!3}6.8 & \cellcolor{gradgreen!25}59.9 & \cellcolor{gradgreen!23}54.2 & 70k \\
    \bottomrule
  \end{tabular*}
\end{table*}

\subsection{RQ1: Can Agents Synthesize a Feasible Itinerary?}
\label{ssec:rq1}

\emph{Largely, no}: even the strongest agent, GPT-5.6, produces a fully-feasible plan on only \textbf{46.2\%} of the 533 solvable tasks (median \textbf{6.6\%}, floor \textbf{0.0\%}; Table~\ref{tab:main_results}). The failure is not spread evenly. Truthfulness, executability, and correct refusal are largely solved at the top---GPT-5.6 is hallucination-free on \textbf{94.9\%} of tasks (Tru), physically executable on \textbf{86.3\%} (Exe), and refuses the impossible correctly on \textbf{97.0\%} (Inf). The wall is \emph{satisfying the traveler}: even GPT-5.6 meets a plan's \emph{full} set of explicit and implicit needs on only \textbf{50.7\%} of tasks (Sat), the category that binds the headline---just above the $46.2\%$ task-perfect rate, while every other category is far higher. A plan is only as usable as its worst requirement group---one unmet need (an ignored persona, a dropped city) fails the user---so a frontier that satisfies the traveler only half the time is much of why no LLM travel planner is yet dependable for everyday use.

The gap is specific to \emph{planning}: refusal is near-saturated (TP-inf $=97.0\%$), so we report the two regimes separately.

\subsection{RQ2: Where Synthesis Breaks, and How the Bottleneck Moves}
\label{ssec:rq2}

\emph{Two dimensions dominate, and the bottleneck migrates with capability.} Because every failure maps to an exact violated dimension, we can decompose \emph{where} agents fail (Appendix Figure~\ref{fig:failure_heatmap}, Table~\ref{tab:failure_decomp}).

\paragraph{Implicit needs (D1) are the universal bottleneck.}
D1 is a \textbf{top-two failure for all 15 agents}, with failure rates from $53.7\%$ to $91.2\%$. It is also the \emph{last} wall: after GPT-5.6 masters every other dimension (next-worst B3 at $13.3\%$), it still fails implicit-need satisfaction on $53.7\%$ of applicable tasks---a four-fold cliff to the next dimension. Satisfying a traveler's \emph{unstated} persona needs, not following explicit instructions, is the frontier of feasible itinerary synthesis.

\paragraph{Spatio-temporal reachability (B3) is the planner/non-planner watershed.}
B3 failure ranges from $13.3\%$ to $93.8\%$---the widest spread of any dimension: whether a day's legs are physically traversable in the time allotted cleanly separates planners from non-planners.

\paragraph{The bottleneck rises with capability.}
Down the ranking (Appendix Table~\ref{tab:failure_decomp}), weak agents fail \emph{everywhere}---Nova-Pro/Qwen3-Next miss D0-key on $78.2\%$/$59.7\%$ of tasks, plus budget, coverage, and scheduling---while the frontier has pushed every mode down except implicit needs, \emph{clearing the easy dimensions first and concentrating the residual failure on the hardest one}. Multi-city coverage is a telling holdout: city \emph{ordering} is nearly solved (D2-single fails $0.4\%$ at the top), yet \emph{coverage} (D2-multi) still fails $6.7\%$--$47.2\%$ and stays a disproportionate drag on task-perfect.

\subsection{RQ3: Does More Deliberation or Compute Help?}
\label{ssec:rq3}

\emph{Not in what our set lets us measure.} \textbf{Reasoning does not help in the one pair we can test}: in the one vendor-matched pair our set permits, Kimi-K2-Thinking reaches only $6.6\%$ TP-feas against Kimi-K2.5's $17.1\%$, with worse failures where controllability matters (D0-key $19.3\%\!\rightarrow\!42.6\%$, B3 $50.7\%\!\rightarrow\!68.5\%$)---a reasoning-vs-controllability tension, though a single-pair, cross-version observation, not a matched ablation. \textbf{And accuracy does not track spend}: GPT-5.6 tops the board at the \emph{lowest} per-task cost among top agents ($46$k tokens), while the weakest are the most expensive (Nova-2-Lite $324$k, DeepSeek-V3.2 $270$k spend five-to-seven times as much for single-digit TP-feas). Cost is a resource axis, not a correctness one, so we hold efficiency out of the correctness headline (cost--quality scatter, Appendix Figure~\ref{fig:cost_quality}).

% =============================================================================
% Section 6: Conclusion
% =============================================================================
\section{Conclusion}

We introduced \textbf{TREK}, a benchmark for \emph{feasible itinerary synthesis}: one plan jointly constraint-correct, hallucination-free, executable, and responsive to a traveler's unstated needs, scored by a deterministic no-judge evaluator against a human-verified gold that scores $1.0$ on all $800$ tasks, so every gap is an agent limitation. Across $15$ agents the strongest is fully feasible on only $46.2\%$ of solvable tasks (median $6.6\%$, floor $0.0\%$): implicit-need satisfaction is the bottleneck and rises with capability, spatio-temporal reachability separates planners from non-planners, and neither reasoning nor extra tokens closes the gap (limitations in Appendix~\ref{app:limitations}).

\bibliographystyle{ACM-Reference-Format}
\bibliography{bib}

\appendix

This supplementary material is organized into six main sections:

\begin{itemize}
    \item \textbf{Appendix~\ref{app:dataset-construction}}: Dataset Construction (\S\ref{app:data-collection}--\ref{app:api-specs}). 
    \item \textbf{Appendix~\ref{app:query-generation}}: Query Design and Related Result (\S\ref{app:impossible-queries}--\ref{app:budget-calibration}).
    \item \textbf{Appendix~\ref{app:benchmark-comparison}}: Benchmark Comparison Details (\S\ref{app:operational-definitions}--\ref{app:per-benchmark-justifications}).
    \item \textbf{Appendix~\ref{sec:exp-details}}: Experimental Details (\S\ref{app:model-specs}--\ref{ssec:implicit-need-understanding}).
    \item \textbf{Appendix~\ref{app:source-reference}}: Source Reference (\S\ref{app:data-layer}--\ref{ssec:visit-durations}).
    \item \textbf{Appendix~\ref{app:limitations}}: Limitations.
\end{itemize}

% =============================================================================
% Appendix Sections
% =============================================================================
% Appendix content for TREK paper
% This file is included via \input{Text/appendix} from main.tex

\section{Dataset Construction}
\label{app:dataset-construction}

This section details the procedures and considerations for building our dataset, encompassing data sourcing, preprocessing, and schema definition.

\subsection{Datasheet, Licensing, and Intended Use}
Following \emph{Datasheets for Datasets}, we summarize provenance and terms.
\textbf{Motivation \& composition.} TREK measures the joint feasibility of LLM travel plans; it holds $800$ query--annotation tasks and a $212{,}530$-record synthetic knowledge base (flights, hotels, attractions, car rentals) over $375$ cities, with $13$ persona tags. It contains \emph{no} personal or user data: every record is produced by deterministic scripts (\S\ref{app:data-collection}), and the only external input is the public-domain OurAirports city/airport scaffold~\cite{ourairports}.
\textbf{Collection.} Fully synthetic and seeded---no scraping, crowd-work, or human subjects; gold references were rated by a 15-annotator panel---10 PhD researchers and 5 travel-industry practitioners---for realism and executability (\S\ref{ssec:qa}; protocol in Appendix~\ref{app:gold-protocol}).
\textbf{Intended use.} Benchmarking deterministic feasibility of tool-using planning agents. \emph{Out of scope}: the synthetic prices and availability must not be used as real travel information, and the persona tags are coarse research constructs, not a taxonomy of real travelers.
\textbf{Distribution \& license.} The synthetic dataset is released under \textbf{CC BY 4.0} and the code under the \textbf{MIT} license; the OurAirports scaffold is public-domain and only city/airport identifiers are reused. Because the entire knowledge base is reproducible from a seed by the released build script, the benchmark is distributed as versioned code plus a checksummed data snapshot.
\textbf{Maintenance.} Corrections re-ship as a new version: the versioned KB rebuilds deterministically from seed, so a data fix propagates to queries, gold, and scores by re-running the pipeline.

\subsection{Gold Verification Protocol}
\label{app:gold-protocol}
The feasible gold itineraries were rated by \textbf{15 annotators}---10 PhD researchers and 5 travel-industry practitioners. Each annotator independently scored $50$ gold plans on a $1$--$5$ scale for two properties jointly: real-world \emph{realism} and \emph{executability} (``could a traveler follow this plan as written?''). The $750$ ratings cover all $533$ feasible gold plans, with roughly $40\%$ of plans multiply-rated for cross-validation; the mean rating was $4.25/5$ ($\approx$\,85\%). Plans that received a low score or drew disagreement between annotators were revised following the annotators' written suggestions and re-checked before release, so the shipped gold reflects this human vetting rather than only scorer self-consistency. The same panel additionally verified template diversity, entity-to-KB mappings, and persona-consistency, with disagreements resolved by discussion; automated scripts independently verified label correctness (via re-scoring), KB back-tracing, and format consistency. This addresses the construct-validity question directly: a human panel---not just the evaluator---judges the reference plans realistic and executable. Human rating of the \emph{agents'} produced plans, and correlation with the automatic per-plan scores, is left to future work (Appendix~\ref{app:limitations}).

\subsection{Data Synthesis}
\label{app:data-collection}
The knowledge base is \emph{synthetic}: every bookable record is produced by deterministic scripts rather than scraped from any live platform, so the dataset relies on no proprietary API and contains no user data. We fix the realistic \emph{schema} of each category (attractions, hotels, rental cars, flights)---the fields a real listing would carry---and then populate values so that they are mutually consistent (prices, coordinates, room types, facilities, and routes all agree). This internal consistency is exactly what makes a deterministic ground truth and an achievable gold possible. The city/airport scaffold is drawn from the public OurAirports dataset~\cite{ourairports}: we retain only cities served by at least one active major airport and having at least one intercity route. The dataset is used solely for research purposes.

\subsection{Per-Category Generation}
\label{app:anonymization}

Each category is produced by a category-specific deterministic procedure. Crucially, we used LLMs only to \emph{design} the generation templates and transformation rules; the actual records were emitted by scripts, eliminating any risk of LLM hallucination in the final data. Because the data is synthetic from the outset, no record corresponds to a real listing, and there is nothing to de-identify.

\paragraph{Flights.}
Flight numbers follow the IATA two-letter carrier-prefix format (e.g., \texttt{AB1234}). Departure and arrival times are sampled within plausible operating hours, and each flight's duration is set consistently with the geographic distance between its endpoints, so that downstream feasibility checks (connection windows, same-day arrivals) are well-defined.

\paragraph{Hotels and Attractions.}
For these two categories, we adopt a \emph{template-based} generation strategy. An LLM was prompted to produce reusable name-generation templates and address-formatting rules (e.g., ``\texttt{[Adjective] [Landmark-type] [City-suffix]}''); these templates were compiled into deterministic scripts that emit names, addresses, and descriptions that are synthetic but realistic. This template-then-script pipeline ensures diversity and consistency without introducing hallucinated factual claims: every name is synthetic by construction, while geographic attributes (city, coordinates) are assigned coherently to preserve spatial reasoning.

\paragraph{Car Rentals.}
Rental-car records are identified by numeric IDs rather than brand names. Vehicle descriptions use standardized category labels (e.g., Economy, SUV, Convertible, EV), and extra services are generic feature tags (e.g., ``GPS navigation,'' ``Child seats''). Pickup locations correspond to airport names from the public OurAirports scaffold~\cite{ourairports}; per-day rates are assigned to realistic ranges for each vehicle category.

\subsection{Symbol Definitions}
\label{app:symbol-def}

Table~\ref{tab:symbol-def} lists the symbols used in our query annotation schema.  Hard constraints ($C_{\text{hard}}$) specify trip-level requirements (group size, budget, duration), service requirements ($\textit{Req}$) enumerate the resource types the agent must book, and implicit preference keywords ($K_{\text{implicit}}$) encode latent traveler personas.

\begin{table}[ht]
    \centering
    \caption{Symbol definitions for query annotations.}
    \label{tab:symbol-def}
    \begin{tabular}{cll}
        \toprule
        \textbf{Symbol} & \textbf{Full Name} & \textbf{Description} \\
        \midrule
        \multicolumn{3}{l}{\textit{Hard Constraints} ($C_{\text{hard}}$)} \\
        $pn$ & \texttt{person\_num} & Number of travelers \\
        $bg$ & \texttt{budget} & Maximum total budget \\
        $dy$ & \texttt{days} & Trip duration in days \\
        \midrule
        \multicolumn{3}{l}{\textit{Service Requirements} ($\textit{Req}$)} \\
        $f$ & \texttt{req\_flight} & Flight booking requirements \\
        $h$ & \texttt{req\_hotel} & Hotel reservation requirements \\
        $c$ & \texttt{req\_car} & Car rental requirements \\
        $a$ & \texttt{req\_attraction} & Attraction visit requirements \\
        \midrule
        \multicolumn{3}{l}{\textit{Implicit Preferences} ($K_{\text{implicit}}$)} \\
        $kw_n$ & \texttt{keyword\_n} & Traveler persona keywords \\
        \bottomrule
    \end{tabular}
\end{table}

\subsection{Constraint Statistics}
\label{app:constraint-stats}

Table~\ref{tab:constraint-stats} reports the per-task constraint counts across all 800 tasks.  Each task contains 8--12 constraints (mean 9.9, of which $1.6$ are implicit persona keywords), comprising three universal hard constraints ($C_{\text{hard}}$), three flight requirements, and a variable number of hotel, car, attraction, and implicit-need specifications. (The Total row below reports the observed per-task min/max; the per-category column extremes never co-occur in a single task, so their sums 7/13 are not attained.)

\begin{table}[ht]
    \centering
    \caption{Per-query constraint statistics (explicit constraints plus the implicit-keyword count).}
    \label{tab:constraint-stats}
    \begin{tabular}{lccc}
        \toprule
        \textbf{Category} & \textbf{Max} & \textbf{Min} & \textbf{Mean} \\
        \midrule
        $C_{\text{hard}}$ (person, budget, days) & 3 & 3 & 3.0 \\
        Flight requirements ($f$) & 3 & 3 & 3.0 \\
        Hotel requirements ($h$) & 2 & 1 & 1.7 \\
        Car requirements ($c$) & 1 & 0 & 0.3 \\
        Attraction requirements ($a$) & 1 & 0 & 0.3 \\
        Implicit keywords ($K_{\text{implicit}}$) & 3 & 0 & 1.6 \\
        \midrule
        \textbf{Total constraints} & \textbf{12} & \textbf{8} & \textbf{9.9} \\
        \bottomrule
    \end{tabular}
\end{table}

\subsection{Knowledge-Base Statistics}
\label{app:kb-stats}
\begin{table}[h]
    \centering
    \small
    \caption{Knowledge-base statistics. All records are synthetic and mutually consistent; 392 airports serve the 375 cities (some cities map to several airports).}
    \label{tab:kb-stats}
    \begin{tabular}{lrr}
        \toprule
        \textbf{Domain} & \textbf{Records} & \textbf{City Coverage} \\
        \midrule
        Flights & 107{,}195 & 375 cities (392 airports) \\
        Hotels & 39{,}396 & 375 cities \\
        Attractions & 55{,}814 & 375 cities \\
        Car Rentals & 10{,}125 & 375 cities \\
        \midrule
        \textbf{Total} & \textbf{212{,}530} & \textbf{375 cities} \\
        \bottomrule
    \end{tabular}
\end{table}

\subsection{Schema Details}
Table~\ref{tab:schema-details} lists the attributes included in each entity type of the knowledge base.

\begin{table}[ht]
  \centering
  \small
  \caption{Schema details for each entity type in the knowledge base.}
  \label{tab:schema-details}
  \renewcommand{\arraystretch}{1.2}
  \begin{tabularx}{\linewidth}{l X}
    \toprule
    \textbf{Entity} & \textbf{Attributes} \\
    \midrule
    Attractions & \texttt{attraction\_id, city\_name, country, type, attraction\_name, overview, ticket\_price, open\_hours, duration\_of\_visit, rate\_of\_restaurant, address, latitude, longitude, facilities\_group, facilities} \\
    Hotels & \texttt{hotel\_id, city\_name, country, name, about, address, latitude, longitude, amenities\_group, amenities, price, rating, star, rate\_of\_restaurant} \\
    Rental Cars & \texttt{car\_id, city\_name, price\_per\_day, pickup\_location, car\_type, capacity, extra\_services, extra\_services\_group} \\
    Flights & \texttt{flight\_id, departure\_city, departure\_airport\_name, arrival\_airport\_name, arrival\_city, departure\_time, arrival\_time, flight\_number, price, departure\_airport\_latitude, departure\_airport\_longitude, arrival\_airport\_latitude, arrival\_airport\_longitude, departure\_airport\_continent, departure\_airport\_country, arrival\_airport\_continent, arrival\_airport\_country, departure\_airport\_iata\_code, arrival\_airport\_iata\_code} \\
    \bottomrule
  \end{tabularx}
\end{table}
\subsection{Facilities and Services}
\label{app:facilities}

We catalogued the most common facilities and services for each entity type by surveying leading online travel platforms and examining combinations favored by different traveler implicit needs segments.  Table~\ref{tab:classic-facilities} lists the core amenities we extracted for attractions, hotels, and vehicle rentals.

\begin{table}[ht]
  \centering
  \renewcommand{\arraystretch}{1.1}
  \setlength{\tabcolsep}{6pt}
  \caption{Classic facilities and services by entity type.}
  \label{tab:classic-facilities}
  \begin{tabularx}{\linewidth}{l X}
    \toprule
    \textbf{Entity}     & \textbf{Core Facilities / Amenities / Extra Services} \\
    \midrule
    Attractions         & Restrooms; Information Desk; Restaurant; Souvenir Shop; First Aid Station; Luggage Storage \\
    Hotels              & Wi-Fi; Luggage Storage; Air Conditioning; Room Cleaning; Safety Deposit Box; Television \\
    Rental Cars         & GPS Navigation; Unlimited Mileage; Insurance Included; Free Cancellation; Air Conditioning; 24/7 Support \\
    \bottomrule
  \end{tabularx}
\end{table}

\subsection{Implicit Needs and Segment-Specific Requirements}
\label{app:implicit-requirements}

Through analysis of travel websites, blogs, and community forums, we distilled thirteen traveler implicit segments characterized by latent preferences.  Table~\ref{tab:implicit-keywords} maps each segment to its keyword, and Tables~\ref{tab:attraction-req},~\ref{tab:hotel-req}, and~\ref{tab:car-req} summarize the supplementary facilities or services required for each segment.

\begin{table}[ht]
  \centering
  \caption{Implicit traveler segments and keywords.}
  \label{tab:implicit-keywords}
  \begin{tabular}{r l}
    \toprule
    \# & \textbf{Segment Keyword} \\
    \midrule
    1  & with children \\
    2  & road trip \\
    3  & elderly travelers \\
    4  & business travelers \\
    5  & with pets \\
    6  & nightlife enthusiast \\
    7  & disabled traveler \\
    8  & fast-paced budget travel \\
    9  & couples trip \\
    10 & solo women \\
    11 & luxury travelers \\
    12 & foodie \\
    13 & photography \\
\bottomrule
  \end{tabular}
\end{table}

\begin{table}[ht]
  \centering
  \renewcommand{\arraystretch}{1.0}
  \setlength{\tabcolsep}{4pt}
  \caption{Attraction facilities per traveler segment.}
  \label{tab:attraction-req}
  \begin{tabularx}{\linewidth}{l X}
    \toprule
    \textbf{Segment}          & \textbf{Additional Facilities} \\
    \midrule
    with children             & family restrooms; nursing rooms \\
    road trip                 & parking \\
    elderly travelers         & wheelchair rental; benches/rest areas \\
    business travelers        & high-speed Wi-Fi \\
    with pets                 & pet-friendly areas; pet water stations; pet rest zones \\
    nightlife enthusiast      & night markets; bars; evening shows \\
    disabled traveler         & ramps; elevators; wheelchair rentals; accessible restrooms \\
    fast-paced budget travel  & city passes; luggage storage; self-guided tours \\
    couples trip              & scenic spots; sunset cruises; couples-only experiences \\
    solo women                & group tours; enhanced security \\
    luxury travelers          & private tours; skip-the-line access; VIP events \\
    foodie                     & food markets; tasting tours \\
    photography               & photo spots; guided photo tours; charging stations \\
    \bottomrule
  \end{tabularx}
\end{table}

\begin{table}[ht]
  \centering
  \caption{Hotel amenities per traveler segment.}
  \label{tab:hotel-req}
  \begin{tabularx}{\linewidth}{l X}
    \toprule
    \textbf{Segment}         & \textbf{Additional Amenities} \\
    \midrule
    with children            & child-friendly rooms; baby cots; stroller storage; diaper tables; nursing rooms \\
    road trip                & parking; 24-hour front desk; EV charging \\
    elderly travelers        & elevators; grab bars; medical contacts; dietary breakfast options \\
    business travelers       & high-speed Wi-Fi; business center; meeting rooms; laundry; printing services \\
    with pets                & pet-friendly rooms; pet rest areas; pet beds; durable flooring \\
    nightlife enthusiast     & 24-hour front desk; late-night room service; on-site bars; rooftop lounges \\
    disabled traveler        & accessible entrances; ramps/lifts; adapted rooms/bathrooms; Braille signage \\
    fast-paced budget travel & free Wi-Fi; communal kitchens; laundry; lockers \\
    couples trip             & bathtubs/jacuzzis; scenic villas; spa access \\
    solo women               & women-only floors; privacy-focused check-in; 24-h security; surveillance; double locks \\
    luxury travelers         & spa; gym; premium bedding; bar \\
    foodie                    & — \\
    photography              & scenic rooms; sunrise calls; photography packages \\
    \bottomrule
  \end{tabularx}
\end{table}

\begin{table}[ht]
  \centering
  \caption{Vehicle services per traveler segment.}
  \label{tab:car-req}
  \begin{tabularx}{\linewidth}{l X}
    \toprule
    \textbf{Segment}         & \textbf{Additional Services} \\
    \midrule
    with children            & child seats; child locks \\
    road trip                & roadside emergency support \\
    elderly travelers        & advanced driving assistance \\
    business travelers       & onboard Wi-Fi \\
    with pets                & pet seat belts; pet-friendly cars \\
    nightlife enthusiast     & late pick-up service \\
    disabled traveler        & wheelchair space; hand controls; accessible vehicles \\
    fast-paced budget travel & — \\
    couples trip             & — \\
    solo women               & women-only cars; secure airport waiting areas \\
    luxury travelers         & first-class cars; chauffeured service; guided-driver packages \\
    foodie                    & — \\
    photography              & remote shooting support \\
    \bottomrule
  \end{tabularx}
\end{table}

\noindent
For full details on data sources and list of travel websites consulted, refer to the “Source Reference” section.  

\subsection{API Specifications}
\label{app:api-specs}

Our evaluation environment exposes \textbf{seven} agent tools: four RESTful search endpoints for domain resources (flights, hotels, attractions, car rentals) and \texttt{submit\_plan} for the final itinerary or typed refusal---the five \emph{billable} tools counted by the efficiency axis---plus two \emph{free} helpers, \texttt{compute\_travel\_time} (returns the exact door-to-door minimum the B3 check uses) and \texttt{write\_note} (a scratchpad whose notes stay in the transcript). The four search endpoints and \texttt{submit\_plan} are specified in Table~\ref{tab:api-specs}.

\begin{table*}[ht]
\centering
\small
\caption{RESTful API endpoint specifications.}
\label{tab:api-specs}
\renewcommand{\arraystretch}{1.15}
\begin{tabularx}{\textwidth}{l l X}
\toprule
\textbf{Tool Name} & \textbf{Type} & \textbf{Parameters} \\
\midrule
\texttt{search\_flights} & Required & \texttt{departure\_city}, \texttt{arrival\_city} (String); \texttt{trip\_type} $\in$ \{\texttt{one\_way}, \texttt{round\_trip}\} \\
 & Optional & \texttt{max\_price} (Float); \texttt{sort\_by} $\in$ \{\texttt{price}, \texttt{departure\_time}\}; \texttt{top\_k} (Int) \\
\midrule
\texttt{search\_hotels} & Required & \texttt{city} (String, comma-separated batch) \\
 & Optional & \texttt{max\_price} (Float); \texttt{min\_star} (Int, 1-5); \texttt{amenity} (String, semantic); \texttt{sort\_by} $\in$ \{\texttt{price}, \texttt{rating}, \texttt{star}\}; \texttt{top\_k} (Int) \\
\midrule
\texttt{search\_attractions} & Required & \texttt{city} (String, comma-separated batch) \\
 & Optional & \texttt{max\_ticket\_price} (Float); \texttt{facility} (String, semantic); \texttt{sort\_by} $\in$ \{\texttt{price}, \texttt{rating}, \texttt{duration\_of\_visit}\}; \texttt{top\_k} (Int) \\
\midrule
\texttt{search\_cars} & Required & \texttt{city} (String, comma-separated batch) \\
 & Optional & \texttt{min\_capacity} (Int); \texttt{max\_price\_per\_day} (Float); \texttt{car\_type} (String); \texttt{extra\_service} (String, semantic); \texttt{top\_k} (Int) \\
\midrule
\texttt{submit\_plan} & Required & \texttt{is\_feasible} (Bool); \texttt{plan} (Object) \\
 & Optional & \texttt{refusal\_reason} (String) \\
\bottomrule
\end{tabularx}
\end{table*}

\section{Query Design and Related Result}
\label{app:query-generation}
Here we describe our methodology for crafting queries, including the typed impossibility classes and the budget-calibration procedure. Our \emph{template-then-populate} design (\S\ref{ssec:query-gen}) deliberately trades linguistic diversity for verifiable grounding: pure templates keep every populated entity, route, and price resolvable against the KB, which is what makes the feasibility labels correct by construction. A natural extension is to paraphrase the populated surface form---e.g., an LLM rewrite that leaves the underlying constraints untouched---to broaden linguistic variety while retaining construction-correct labels.

\subsection{Impossible Queries}
\label{app:impossible-queries}
A third of the benchmark ($267$ tasks) is \emph{provably infeasible} by construction, split evenly across three typed causes ($89$ each), so that a correct agent must not only refuse but name the right reason (scored by D4, \S\ref{ssec:scoring-dimensions}):
\begin{itemize}
  \item \textit{Nonexistent Entity}: the query mandates a named hotel, attraction, or car type absent from the knowledge base, so no compliant plan can book it.
  \item \textit{No Flight Route}: no flight connects a required pair of cities in the knowledge base.
  \item \textit{Insufficient Budget}: the budget is set strictly below the achievable cost floor (Appendix~\ref{app:budget-calibration}), so no plan fits.
\end{itemize}
\noindent Each class is generated deterministically and re-verified by the evaluator before shipping. \textbf{Example (No Flight Route):}
\begin{quote}
Could you plan a round-trip expedition from Addis Ababa to Fort-de-France for disabled traveler, foodie, and photography? I’ll stay 3 days, need 2 rooms, and my budget is 8400 USD.
\end{quote}
We flag the infeasibility via our deterministic entity-, route-, and budget-verification procedures. These three causes are chosen for \emph{unambiguous} machine-checkability and form an extensible starting point rather than a closed set: opening-hour conflicts, resource-capacity limits, and cross-domain incompatibilities are natural additions that preserve the same typed, deterministically verifiable structure.

\subsection{Conflict of Implicit Need Keywords}
An earlier hand-written set of \emph{nineteen} ``incompatible'' persona pairs (Table~\ref{tab:retired-pairs}) was \textbf{retired}. An audit of all $105{,}335$ knowledge-base resources found both members of every such pair co-occurring on $2{,}327$--$6{,}269$ resources each---against a $5{,}529$-resource median for the pairs the \emph{same} rules called legal---so the asserted conflicts were statistically indistinguishable from the permitted ones and were not actually unsatisfiable. Several also encoded age-, disability-, or gender-based stereotypes (e.g., \emph{elderly travelers}\,$\times$\,\emph{solo women}, \emph{disabled traveler}\,$\times$\,\emph{nightlife enthusiast}), and the set was internally inconsistent---it forbade \emph{couples trip}\,$\times$\,\emph{solo women} yet allowed \emph{couples trip}\,$\times$\,\emph{with children}. Treating these as hard contradictions would have penalized agents for correct plans and baked bias into the ground truth.

\begin{table}[h]
  \centering \footnotesize
  \setlength{\tabcolsep}{4pt}
  \caption{The \emph{nineteen} retired ``incompatible'' persona pairs. $\dagger$: encoded an age/disability/gender stereotype; $\ddagger$: still excluded, but under a principled rule (party-size, or the single same-axis spend conflict). All others are now treated as jointly satisfiable and scored on the merits by D1.}
  \label{tab:retired-pairs}
  \begin{tabular}{@{}l@{}}
    \toprule
    \textit{with children} $\times$ \{nightlife, fast-paced budget, business, solo women\} \\
    \textit{elderly travelers} $\times$ \{nightlife$\dagger$, fast-paced budget, solo women$\dagger$\} \\
    \textit{business travelers} $\times$ \{with pets, fast-paced budget, couples, photography\} \\
    \textit{with pets} $\times$ \{nightlife, fast-paced budget, luxury, foodie\} \\
    \textit{nightlife enthusiast} $\times$ \textit{disabled traveler}$\dagger$ \\
    \textit{disabled traveler} $\times$ \textit{fast-paced budget travel}$\dagger$ \\
    \textit{fast-paced budget travel} $\times$ \textit{luxury travelers}$\ddagger$ \\
    \textit{couples trip} $\times$ \textit{solo women}$\ddagger$ \\
    \bottomrule
  \end{tabular}
\end{table}

The current generator marks a persona combination infeasible only when it is \emph{logically} contradictory:
\begin{itemize}[leftmargin=*, itemsep=2pt]
  \item \textbf{Party-size contradictions} derived from the stated traveler count (e.g., a ``solo'' persona with a party of four).
  \item The single genuine same-axis conflict, \emph{fast-paced budget travel} vs.\ \emph{luxury travelers}, which are mutually exclusive service tiers.
\end{itemize}
\noindent All other persona combinations are treated as jointly satisfiable and are scored on the merits by D1 (\S\ref{ssec:scoring-dimensions}).

\subsection{Budget Calibration}
\label{app:budget-calibration}

Budgets are set \textbf{by construction} so that each task's feasibility is guaranteed at generation time rather than annotated afterward, with no circular dependency on model behavior. The generator prices a reference plan with the \emph{same} cost model the scorer bills with, so a budget is always defined relative to the true achievable cost of the task.

\paragraph{Procedure.}
For each task, a deterministic, seeded script (reading only the versioned KB; no LLM) computes two cost anchors: $C_{\min}$, the cost of the cheapest constraint-compliant reference plan, and $C_{\text{floor}}$, an achievable cost floor. The budget is then set relative to these anchors according to the task's intended class (Algorithm~\ref{alg:budget}):

\begin{itemize}[leftmargin=*, itemsep=2pt]
    \item \textbf{Feasible (tight):} $b = \lceil C_{\min}\cdot u\rceil$ with $u \in [1.02, 1.10]$---just above the cheapest valid plan.
    \item \textbf{Feasible (loose):} $b = \lceil C_{\min}\cdot u\rceil$ with $u \in [1.35, 1.80]$---comfortably above.
    \item \textbf{Budget-infeasible:} $b = \lfloor C_{\text{floor}}\cdot u\rfloor$ with $u \in [0.55, 0.92]$, capped strictly below $C_{\text{floor}}$---so no plan can fit.
\end{itemize}

\begin{algorithm}[h]
\caption{By-Construction Budget Assignment}\label{alg:budget}
\begin{algorithmic}[1]
\Require Task $q_i$; knowledge base $\mathcal{KB}$; intended class; seeded RNG
\Ensure Budget $b_i$ with a feasibility label correct by construction
\State $C_{\min} \gets$ cost of the cheapest constraint-compliant plan for $q_i$ \Comment{same cost model as the scorer}
\State $C_{\text{floor}} \gets$ achievable cost floor for $q_i$
\If{class $=$ tight} $\; b_i \gets \lceil C_{\min}\cdot \mathrm{U}(1.02, 1.10)\rceil$
\ElsIf{class $=$ loose} $\; b_i \gets \lceil C_{\min}\cdot \mathrm{U}(1.35, 1.80)\rceil$
\Else{} $\; b_i \gets \min\!\bigl(\lfloor C_{\text{floor}}\cdot \mathrm{U}(0.55, 0.92)\rfloor,\; C_{\text{floor}} - \max(50,\, 0.05\,C_{\text{floor}})\bigr)$
\EndIf
\State re-score $q_i$ at $b_i$ with the authoritative evaluator; keep iff realized label $=$ intended
\end{algorithmic}
\end{algorithm}

\paragraph{Correct by construction.}
Every generated task is re-scored by the authoritative evaluator (\S\ref{sec:evaluation}) before it ships, and only tasks whose realized feasibility matches the intended class are kept; the $89$ budget-infeasible tasks all satisfy $b_i < C_{\text{floor}}$. Generation is fully deterministic (a seeded RNG over the fixed KB), so the entire task set is reproducible from source. Because the budget is defined by the same cost model the scorer uses, it cannot encode a model-specific preference---a feasible task's budget clears the cheapest valid plan, while a budget-infeasible task's budget sits below the achievable floor. Across the $533$ feasible tasks the tight and loose bands are near-evenly split ($267$ tight, $266$ loose; mean slack $b_i/C_{\min}=1.31$, median $1.10$, range $1.02$--$1.80$), so about half the feasible tasks leave an agent little headroom before overspending---a deliberate stress on budget adherence---while the other half are comfortable.

\section{Benchmark Comparison Details}
\label{app:benchmark-comparison}

This section provides operational definitions for the differentiating capability columns in Table~\ref{tab:benchmarks} and concrete per-benchmark justifications.

\subsection{Operational Definitions}
\label{app:operational-definitions}
\begin{itemize}[leftmargin=*, itemsep=2pt]
    \item \textbf{Scoring}: rule-based (\emph{Rules}, no LLM judge), LLM-structuring-then-rules (\emph{Ext.+Rules}), executable \emph{DSL}, simulation (\emph{Sim.}), execution (\emph{Exec.}), exact \emph{Match}, or LLM \emph{Judge}. DSL and Sim.\ are also judge-free; TREK's differentiator is the \emph{combination} of properties in this table, not determinism alone.
    \item \textbf{API (typed tool sandbox)}: the agent acts through structured tool calls with HTTP-level parameter validation, typed JSON schemas, and structured error responses, as opposed to free-form function calls or browser actions.
    \item \textbf{Gold\,$=$\,1}: a per-task reference solution is provided \emph{and} shown to attain the evaluator maximum on every task (not merely provided).
    \item \textbf{Impl. (implicit preference)}: whether implicit persona preferences (e.g., luxury, foodie) are scored, and \emph{how}---\emph{Det.} (deterministically, e.g., TREK's facility set-intersection against resource metadata) or \emph{Judge} (via an LLM rubric).
    \item \textbf{Infeas. (infeasibility detection)}: the benchmark includes annotated infeasible queries where the correct behavior is to refuse; \emph{Typed} means the cause is a machine-checkable route/entity/budget proof, \emph{Untyped} a descriptive category only.
    \item \textbf{Effic. (efficiency)}: an explicit efficiency metric penalizes redundant tool calls against an oracle minimum, jointly assessed with planning quality.
\end{itemize}

\subsection{Per-Benchmark Justifications}
\label{app:per-benchmark-justifications}

\paragraph{TravelPlanner~\cite{travelplanner}.}
Queries specify budgets, date ranges, and group sizes that must be jointly satisfied, but it invokes tools via Python functions without HTTP parameter validation or typed error responses (API\,\xmark); it contains no annotated infeasible queries---all tasks are designed to be solvable (Infeas\,\xmark); it does not model implicit persona preferences (Impl\,\xmark); and it measures only final-plan correctness without a tool-call efficiency metric (Effic\,\xmark).

\paragraph{TravelBench (Cheng et al.)~\cite{travelbench_acl}.}
A multi-turn, tool-using travel benchmark over real Amap map/navigation APIs. Its sandbox validates tool calls against typed specifications---invalid tool name, missing required arguments, type mismatch, or other schema violations---so it provides a typed tool sandbox (API\,\cmark); it elicits implicit preferences from de-identified user profiles through multi-turn interaction (Impl\,\cmark); and it annotates infeasible requests with descriptive categories (missing-info, missing-tool, no-actionable-intent) rather than machine-checkable typed proofs (Infeas: Untyped). However, it is scored by an LLM-as-judge rubric with a meta-judge (no per-task reference shown maximal, Gold\,\xmark); and its ``tool-use penalty'' is a tool-call \emph{error} rate ($1-\text{erroneous}/\text{total calls}$), which charges malformed calls but not redundant valid ones, so it is not an efficiency-versus-oracle metric (Effic\,\xmark).

\paragraph{ToolBench~\cite{toolllm}.}
Evaluates 16k+ real REST APIs with parameter schemas (API\,\cmark) and includes unsolvable queries that agents should recognize, though as descriptive categories rather than typed proofs (Infeas: Untyped). Its ToolEval reports only pass and win rates, with no separately scored tool-use efficiency metric (Effic\,\xmark). Moreover, its queries target individual API tasks without joint budget/temporal/capacity trade-offs, and it has no persona or facility-matching mechanism (Impl\,\xmark).

\paragraph{WebArena~\cite{zhou2023webarena}.}
Evaluates agents on realistic websites; some tasks are inherently impossible, requiring the agent to recognize and report failure---untyped infeasibility (Infeas: Untyped). However, agents interact via browser actions (clicks, form fills) rather than structured API calls with typed schemas (API\,\xmark); no implicit preference scoring exists (Impl\,\xmark); and success is a single end-state check without efficiency scoring (Effic\,\xmark).

\section{Experimental Details and Results}
\label{sec:exp-details}

\subsection{Model Specifications}
\label{app:model-specs}

Table~\ref{tab:model-specs} lists all 15 evaluated models, served through Amazon Bedrock via one of three request dialects: the Converse (\emph{runtime}) API, an OpenAI-compatible chat (\emph{mantle}) API, and the OpenAI Responses API (which GPT-5.x requires). All models use temperature $0.0$; \texttt{max\_tokens} is set generously per model (mostly $32{,}768$) so that truncation is rare---a departure from earlier single-global-limit setups. GPT-5.6 is the sole frontier closed model; Kimi-K2-Thinking is the one explicit reasoning variant. The Claude family is geo-blocked from the run location at the provider level and could not be evaluated.

\noindent Our run transcripts record a per-task \texttt{truncation\_events} count, letting us rule out truncation as a confound. An intermediate turn hit the token limit on $0$ tasks for $11$ of the $15$ models, and---critically for the two models capped at $8{,}192$ tokens (Llama-4-Maverick, Nova-Pro)---on only $1$ and $4$ tasks respectively. Those two models' bottom placement reflects \emph{stalls} (no submission within the call budget---Nova-Pro's four truncated tasks all stalled without submitting) and malformed submissions, not token truncation. The remaining two non-zero models are Gemma-4-31B ($13$ tasks) and GPT-OSS-20B ($113$ tasks), both $32{,}768$-token models on which truncation rarely blocked the final submission---all $13$ of Gemma-4-31B's truncated tasks and $110$ of GPT-OSS-20B's $113$ still submitted a plan---so token limits do not explain their scores.

\begin{table*}[ht]
  \centering
  \small
  \caption{Evaluated models: vendor, exact Bedrock model identifier, request dialect, and per-model token limit. All run at temperature $0.0$ in region \texttt{us-east-1}. The identifiers are reproduced verbatim from \texttt{trek\_models.json} in the released artifact, which is the authoritative record of what was invoked; a display name alone (``Nova-Pro'') does not pin a specific served model, whereas the identifier does.}
  \label{tab:model-specs}
  \resizebox{\textwidth}{!}{%
  \begin{tabular}{@{}llllr@{}}
    \toprule
    \textbf{Model} & \textbf{Vendor} & \textbf{Bedrock model ID} & \textbf{API} & \textbf{\texttt{max\_tokens}} \\
    \midrule
    GPT-5.6~\cite{model_gpt56} & OpenAI & \texttt{openai.gpt-5.6-sol} & Responses & 32{,}768 \\
    Kimi-K2.5~\cite{model_kimik25} & Moonshot & \texttt{moonshotai.kimi-k2.5} & Converse & 32{,}768 \\
    Kimi-K2-Thinking~\cite{model_kimik2thinking} & Moonshot & \texttt{moonshot.kimi-k2-thinking} & Converse & 32{,}768 \\
    GLM-5~\cite{model_glm5} & Zhipu & \texttt{zai.glm-5} & Converse & 32{,}768 \\
    GLM-4.7~\cite{model_glm47} & Zhipu & \texttt{zai.glm-4.7} & Converse & 32{,}768 \\
    Gemma-4-31B~\cite{model_gemma4} & Google & \texttt{google.gemma-4-31b} & Mantle & 32{,}768 \\
    Grok-4.3~\cite{model_grok43} & xAI & \texttt{xai.grok-4.3} & Mantle & 32{,}768 \\
    Mistral-Large-3~\cite{model_mistrallarge3} & Mistral & \texttt{mistral.mistral-large-3-675b-instruct} & Converse & 32{,}768 \\
    DeepSeek-V3.2~\cite{model_deepseekv32} & DeepSeek & \texttt{deepseek.v3.2} & Converse & 32{,}768 \\
    GPT-OSS-120B~\cite{model_gptoss} & OpenAI & \texttt{openai.gpt-oss-120b-1:0} & Converse & 32{,}768 \\
    GPT-OSS-20B~\cite{model_gptoss} & OpenAI & \texttt{openai.gpt-oss-20b-1:0} & Converse & 32{,}768 \\
    Qwen3-Next-80B~\cite{model_qwen3next80b} & Alibaba & \texttt{qwen.qwen3-next-80b-a3b} & Converse & 32{,}768 \\
    Llama-4-Maverick~\cite{model_llama4maverick} & Meta & \texttt{us.meta.llama4-maverick-17b-instruct-v1:0} & Converse & 8{,}192 \\
    Nova-Pro~\cite{model_novapro} & Amazon & \texttt{us.amazon.nova-pro-v1:0} & Converse & 8{,}192 \\
    Nova-2-Lite~\cite{model_nova2lite} & Amazon & \texttt{us.amazon.nova-2-lite-v1:0} & Converse & 32{,}768 \\
    \bottomrule
  \end{tabular}%
  }
\end{table*}

\subsection{Implementation Details}
\label{app:implementation}

\paragraph{Working memory (no response cache).}
The agent keeps a \emph{lossless} working transcript: every tool result is retained verbatim in the conversation and re-sent to the model each turn, so a model never has to reconstruct a value it already saw. We deliberately do \emph{not} cache or deduplicate tool responses, so token cost grows with the number of calls---a cost the efficiency axis (\S\ref{app:scoring-details}) accounts for.

\paragraph{Resume.}
Results are appended per task and flushed to disk immediately; an interrupted run resumes by skipping any \texttt{query\_index} already present in the output file, so no completed task is re-run and no fixed checkpoint interval is needed.

\paragraph{Retry.}
Each provider call retries up to five times with exponential backoff ($2^{a}$ seconds at attempt $a=0,\dots,4$) before the task is recorded as failed, handling transient provider/network errors.

\paragraph{Pipeline Architecture.}
The evaluation pipeline serves all models through Amazon Bedrock via a unified interface spanning three request dialects (Converse, OpenAI-compatible chat, and the Responses API) with a common request/response schema. Queries are processed in parallel with configurable concurrency limits. All evaluation metrics are computed by the same deterministic constraint checker across all runs, eliminating scorer variance. All experiments use temperature 0.0.

\subsection{Scoring Details}
\label{app:scoring-details}

\paragraph{Explicit Constraint Checklist.}
We verify the following constraint types against the generated plan:
\begin{itemize}[leftmargin=*]
  \item \textbf{Trip Structure}: Departure city, arrival city(s), trip type (one-way/round-trip)
  \item \textbf{Hotel Specifications}: Number of rooms, star rating, price constraints
  \item \textbf{Car Rental}: Vehicle type, capacity requirements, daily price limits
  \item \textbf{Attractions}: Named attractions that must be visited
  \item \textbf{Budget}: Total cost must not exceed the specified budget
\end{itemize}

\paragraph{Implicit Need Scoring.}
Persona needs are scored by \emph{deterministic facility set-intersection}, with no embeddings and no similarity threshold. Each persona $p$ maps (Appendix~\ref{app:implicit-requirements}) to a fixed set of required facilities $F_p$. For each applicable (persona, resource-type, stay-city) cell, the cell scores $1$ if the best booked resource of that type in that city carries \emph{any} facility in $F_p$ (the intersection is nonempty) and $0$ otherwise; the D1 score is the mean over cells. The \emph{luxury} and \emph{foodie} personas add a quality test \emph{on top of} the facility set rather than replacing it: a luxury hotel cell averages a premium-amenity hit (Spa / Gym / premium bedding / bar) with a star $\geq 5$ check---so a $5$-star hotel carrying none of those amenities scores $0.5$, not $1$---while foodie requires restaurant rating $\geq 4.0$ on hotels and a food-facility set (food markets, tasting tours) on attractions. A resource that does not resolve to a knowledge-base record contributes zero, so fabricated amenities cannot earn credit. Because every check is a deterministic set or threshold test over KB fields, D1 is bit-reproducible and has no embedding to tune. One cell is excluded as non-discriminative---the (\emph{fast-paced budget travel}, attraction) pair, which nearly all KB attractions satisfy and which therefore carries no signal; this is the sole KB-statistical exception to the otherwise task-derived applicability.

\paragraph{D0-src: Verification Field Specification.}
Each entity in the plan is matched against the knowledge base using core identifier fields only:

{\footnotesize
\begin{tabular}{@{}lp{3.2cm}l@{}}
\toprule
\textbf{Entity} & \textbf{Verified fields} & \textbf{Match} \\
\midrule
Flight & \texttt{flight\_no}, \texttt{dep\_city}, \texttt{arr\_city}, schedule & exact (AND) \\
Hotel & \texttt{name}, \texttt{city} & case-insens. \\
Attraction & \texttt{name}, \texttt{city} & case-insens. \\
Car & \texttt{car\_type}, \texttt{city} & case-insens. \\
\bottomrule
\end{tabular}
}

\noindent\textbf{Binary and zero-tolerance.}
D0-src is $1.0$ iff \emph{every} plan entity is verified ($n_v = n_e$) and $0.0$ otherwise, where $n_v$ counts verified entities and $n_e$ the total. An entity is verified only if its core identifiers match a KB record---for flights, the number, endpoints, \emph{and} a timetable-consistent schedule; for cars, type-with-city---and a city-less or otherwise unresolvable named entity counts as a hallucination. A single unverifiable entity therefore zeroes the task. We deliberately reject a proportional $n_v/n_e$ score: as a graded ratio, one fabricated hotel in a large itinerary costs only a few percent, letting a plan hide a hallucination behind otherwise-correct entities---whereas a deployable plan with one invented entity is simply wrong. The per-entity verified fraction is retained only as a severity diagnostic.

\noindent\textbf{Why identifier-only verification.}
We verify core identifiers (name, city, flight number, route) rather than mutable attributes (price, star rating, operating hours) for two reasons: (1)~prices and ratings can be legitimately approximated or rounded by agents during planning; and (2)~the primary goal of D0-src is to detect \emph{entity existence hallucinations}---fabricated hotels, flights, or attractions that do not exist in the knowledge base---not minor attribute discrepancies. Attribute accuracy (e.g., correct price) is instead captured by the cost computation in D3 (Budget Adherence), where KB prices are used as ground truth.

\paragraph{Haversine Distance Formula.}
For temporal feasibility scoring, we compute distances between consecutive locations using:
\begin{equation}
d = 2R \cdot \arcsin\left(\sqrt{\sin^2\left(\frac{\phi_2-\phi_1}{2}\right) + \cos(\phi_1)\cos(\phi_2)\sin^2\left(\frac{\lambda_2-\lambda_1}{2}\right)}\right)
\end{equation}
where $R = 6371$ km is Earth's radius, $(\phi_1, \lambda_1)$ and $(\phi_2, \lambda_2)$ are the latitude/longitude pairs in radians. Given $d$, the \emph{minimum door-to-door travel time} $t_{\min}$ (in minutes) is the faster of two modes:

\begin{equation}
    t_{\min}(d) = \min\bigl(\; \underbrace{\max(45,\; 15 + d)}_{\text{surface}},\;\; \underbrace{180 + 60\,d/700}_{\text{air}} \;\bigr)
    \label{eq:travel-time}
\end{equation}

\paragraph{B2: Opening-Hours Compliance.}
A scheduled attraction visit is \emph{compliant} iff it resolves to a KB record, its $[\textit{visit\_start},\textit{visit\_end}]$ interval lies inside that record's opening hours, and it covers the required dwell time---taken as the shorter of the KB minimum visit duration and $60$ minutes. The $60$-minute cap is deliberate: enforcing the full KB minimum would fail the human-verified gold on $26\%$ of its timed visits, so the shorter of the two is required. A missing or unparseable \textit{visit\_start}, or a visit whose entity is absent from the KB, counts as a violation rather than a free pass (so omitting or fabricating times cannot inflate the score). The B2 score is the number of compliant visits divided by the larger of the scheduled-visit count and the stay-city count; the stay-city floor in the denominator prevents a plan from maximizing B2 by scheduling almost nothing, and is gold-safe because the gold schedules at least one visit per stay city.

\begin{figure}[t]
  \centering
  \includegraphics[width=0.86\columnwidth]{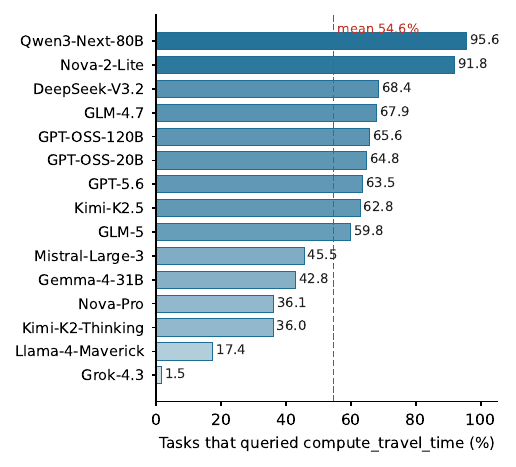}
  \caption{Per-model use of the free \texttt{compute\_travel\_time} helper: the fraction of the $800$ tasks on which the agent queried the exact door-to-door travel-time model that B3 scores against. All $15$ models used it; the mean is $54.6\%$ (dashed). Usage did not guarantee feasibility---Qwen3-Next queried it most yet fails B3 most---so B3 failures are a scheduling gap, not an unqueryable rule.}
  \label{fig:ctt_usage}
\end{figure}

\paragraph{B3: Travel-Time Model.}
Surface transit covers $d$\,km at $60$\,km/h with a $15$-minute buffer and a $45$-minute floor; air travel adds $180$ minutes of end-to-end airport overhead (transfer, check-in, security, baggage) to a $700$\,km/h cruise. Taking the \emph{minimum} over modes makes $t_{\min}$ monotone non-decreasing in distance---surface wins below ${\sim}180$\,km, air above---repairing an earlier piecewise form that charged more time at $300$\,km than beyond it and thus let a distance-blind constant-gap schedule pass. Crucially, this is the \emph{same} function exposed to the agent through the \texttt{compute\_travel\_time} helper (\S\ref{ssec:exp-setup}), so no plan is scored against a gap it could not have looked up. All $15$ models invoked this helper, which was queried on $54.6\%$ of task attempts overall (Figure~\ref{fig:ctt_usage}; from Grok-4.3's $1.5\%$ to Qwen3-Next-80B's $95.6\%$). Heavy use did not rescue scheduling---Qwen3-Next queried the exact rule on $95.6\%$ of tasks yet still failed B3 on $93.8\%$---confirming that B3 failures reflect a genuine spatio-temporal scheduling gap, not a rule the agent could not access. Consecutive events are evaluated \textbf{within each day only} (overnight stays provide ample transfer time) and are sorted by start time first; a booking that must be scheduled but carries no parseable time counts as a \emph{violation} rather than being skipped, so omitting times cannot make the dimension disappear. The score is $1 - v/n$, where $v$ is the number of violated transitions and $n$ the number of within-day consecutive-event pairs.

\paragraph{D0-key vs. D2 Boundary.}
D0-key and D2 operate at different granularities and answer different questions:

\begin{itemize}[leftmargin=*, itemsep=2pt]
    \item \textbf{D0-key (entity-level)}: ``Does the plan contain the specific entity requested?'' Checked items include departure/arrival cities \emph{on individual flight tickets}, named hotels, car types, named attractions, and daily-car presence for road trips. Each item is verified by string matching against the corresponding structured field.
    \item \textbf{D2 (set-level)}: ``Does the plan visit \emph{all} required destination cities?'' \textbf{All-or-nothing}: $1.0$ iff $\text{required} \subseteq \text{visited}$ and $0.0$ otherwise, where \emph{visited} is the set of cities the plan actually books (hotel/attraction/car cities and flight-arrival cities).
\end{itemize}

\noindent\textbf{Why both are necessary.} Consider a query requiring travel to Paris and London, staying at ``Grand Hyatt Paris.'' A plan that visits both cities but books the wrong hotel scores D2\,=\,1.0 but loses points on D0-key. Conversely, a plan that books the Grand Hyatt Paris but omits London entirely scores the hotel constraint in D0-key but scores D2\,=\,0.0 (a required city is missing). The two dimensions are complementary: D0-key captures fine-grained entity compliance, while D2 captures macro-level itinerary structure.

\noindent\textbf{Why D0-key's city items do not double-count with D2.} In D0-key, the arrival-city check verifies that a \emph{flight ticket's arrival field} matches the requested city---it measures whether the agent correctly booked a flight \emph{to} that city, not whether the city appears anywhere in the plan. D2, by contrast, checks the set of cities the plan \emph{books} (hotel/attraction/car cities and flight-arrival cities), sharing the flight-arrival field with D0-key but at set rather than per-ticket granularity.

\noindent\textbf{D2-S significance.} Single-city queries yield D2-S\,$\neq$\,1.0 when agents hallucinate an incorrect destination or swap departure and arrival cities---failure modes observed in up to $25\%$ of single-city queries for the weakest models, and under $1\%$ for the strongest (Table~\ref{tab:failure_decomp}).

\noindent\textbf{D2 single vs.\ multi.} Each query is either single-city or multi-city, so exactly one of D2-S and D2-M applies per query; whichever applies enters the Satisfaction conjunction as a single all-or-nothing check (all required cities booked, or fail). We report the two subgroups separately (Table~\ref{tab:failure_decomp}) so that single- and multi-city coverage remain visible independently rather than blended into one number.

\paragraph{Efficiency: Oracle Minimum Calls and Cost Penalty.}
The oracle minimum $\text{min\_calls}$ is the theoretical lower bound on tool invocations needed to gather all required information and submit a plan. It is computed per task, where $k$ denotes the number of stay cities:

\begin{small}
\begin{tabular}{@{}llp{3.4cm}@{}}
\toprule
\textbf{Resource} & \textbf{Min calls} & \textbf{Rationale} \\
\midrule
Flights & $n$ (inter-city legs) & One search per required leg \\
Hotels & $k$ if required, else 0 & One search per stay city \\
Attractions & $k$ if required, else 0 & One search per stay city \\
Cars & $k$ if required, else 0 & One search per stay city \\
Submit & 1 & One \texttt{submit\_plan} call \\
\midrule
Total & $\max(\sum, 2)$ & Floor ensures $\geq$1 search + 1 submit \\
\bottomrule
\end{tabular}
\end{small}

\noindent The efficiency score is \emph{two-sided} in the call count and additionally penalizes token overrun:
\[
    \text{Eff} = \underbrace{\max\!\Bigl(0,\, 1 - \tfrac{\gamma}{100}\,\text{over}\Bigr)}_{\text{over-calling}} \cdot \underbrace{\max\!\Bigl(0,\, 1 - \tfrac{\text{under}}{\text{min\_calls}}\Bigr)}_{\text{under-calling}} \cdot \underbrace{e^{-\beta_\tau \max(0,\, T_{\text{tok}}/B - 1)}}_{\text{token overrun}}
\]
where $\text{over} = \max(0, \text{actual} - \text{min\_calls})$, $\text{under} = \max(0, \text{min\_calls} - \text{actual})$, $\gamma = 150/C_{\max} = 10$ points per surplus call ($C_{\max} = 15$ is the hard cap), $T_{\text{tok}}$ is the total tokens spent, and $B$ is a task-derived token budget that grows with $\text{min\_calls}$ (generous, meant to catch a model that loops for dozens of turns, not normal operation).

\noindent\textbf{Why efficiency is two-sided.} Both over- and under-calling are penalized, because efficiency means \emph{reaching the goal} with few calls, not merely making few calls. A plan that skips required searches cannot have grounded its bookings, so under-calling is charged in proportion to the work it omitted ($\text{under}/\text{min\_calls}$) rather than rewarded for brevity; an earlier one-sided form gave a 3-city task answered with half the required searches a perfect efficiency score. Because efficiency is a resource-cost proxy rather than a correctness requirement, it is reported as a standalone axis and is \emph{not} folded into the conjunctive headline (\S\ref{ssec:composite-scores}).

\paragraph{Category and Overall Aggregation.}
The composite is the \emph{geometric} mean of the four correctness categories (efficiency is held out; see \S\ref{ssec:composite-scores}):
\begin{equation}
    \text{Overall} = \bigl(\bar{C}_{\text{sat}}\cdot C_{\text{truth}}\cdot \bar{C}_{\text{exec}}\cdot \bar{C}_{\text{infeas}}\bigr)^{1/4}
    \label{eq:overall-geomean}
\end{equation}
\noindent where each category is \textbf{all-or-nothing}: on a task it counts as met only if \emph{every} applicable dimension in it equals $1.0$, and the category score is the fraction of applicable tasks that qualify (a dimension is skipped where inapplicable, e.g., D1 on tasks without a persona):
\begin{itemize}[leftmargin=*, itemsep=1pt]
    \item $\bar{C}_{\text{sat}}$: fraction of tasks with $\text{D0-key}{=}\text{D1}{=}\text{D2}{=}\text{D3}{=}1$ (Constraint Satisfaction).
    \item $C_{\text{truth}} = \text{D0-src}$ (Truthfulness; a single all-or-nothing dimension).
    \item $\bar{C}_{\text{exec}}$: fraction of tasks with $\text{B2}{=}\text{B3}{=}1$ (Executability).
    \item $\bar{C}_{\text{infeas}} = \text{D4}$ over the 267 infeasible tasks (Infeasibility handling; equals the infeasible task-perfect rate).
\end{itemize}
\noindent Efficiency is reported as a separate cost axis and is \emph{not} folded into this composite. The four category scores appear in Table~\ref{tab:main_results}; a single zero category is floored at a small $\epsilon$ before the product.

\subsection{System Prompt}
\label{app:system-prompt}

Our evaluation framework consists of two core components: (1) \textbf{Tool Calling} for resource retrieval, and (2) \textbf{Planning} for itinerary generation. All models receive identical system instructions to ensure fair comparison.

\subsubsection{Full System Prompt and Tool Schemas}

The following content is the exact system prompt provided to the agents, including the full JSON schemas for all tools.

\begin{tcolorbox}[colback=gray!8, colframe=gray!40, title={\small\bfseries Complete System Prompt}, fonttitle=\sffamily, breakable, sharp corners, boxrule=0.5pt, left=4pt, right=4pt, top=2pt, bottom=2pt]
\begin{scriptsize}\ttfamily
Tools are invoked through Bedrock-native function calling: the agent emits structured tool-use blocks directly (there is no ReAct \texttt{thought}/\texttt{action} text protocol). The seven tool schemas below are passed to every model verbatim as the Bedrock \texttt{toolConfig}; the complete system prompt follows.

\#\# Tool schemas (Bedrock \texttt{toolConfig})\\
\{\\
\quad "search\_flights": \{\\
\qquad "name": "search\_flights",\\
\qquad "description": "Search for flights between two cities.",\\
\qquad "parameters": \{\\
\qquad\quad "type": "object",\\
\qquad\quad "properties": \{\\
\qquad\qquad "departure\_city": \{"type": "string", "description": "The city to depart from"\},\\
\qquad\qquad "arrival\_city": \{"type": "string", "description": "The destination city"\},\\
\qquad\qquad "trip\_type": \{"type": "string", "enum": ["one\_way", "round\_trip"], "description": "Type of trip"\},\\
\qquad\qquad "max\_price": \{"type": "number", "description": "Maximum price filter"\},\\
\qquad\qquad "sort\_by": \{"type": "string", "enum": ["price", "departure\_time"], "description": "Sort results by this field"\},\\
\qquad\qquad "top\_k": \{"type": "integer", "description": "Number of results to return (default 10)"\}\\
\qquad\quad \},\\
\qquad\quad "required": ["departure\_city", "arrival\_city", "trip\_type"]\\
\qquad \}\\
\quad \},\\
\quad "search\_hotels": \{\\
\qquad "name": "search\_hotels",\\
\qquad "description": "Search for hotels in one or multiple cities. Use comma-separated cities for batch query.",\\
\qquad "parameters": \{\\
\qquad\quad "type": "object",\\
\qquad\quad "properties": \{\\
\qquad\qquad "city": \{"type": "string", "description": "City or comma-separated cities (e.g., 'Paris' or 'Paris,London,Tokyo')"\},\\
\qquad\qquad "name": \{"type": "string", "description": "Look up a SPECIFIC hotel by name -- the only reliable way to confirm an exact hotel exists, since a browse returns only top\_k rows"\},\\
\qquad\qquad "max\_price": \{"type": "number", "description": "Maximum price per night"\},\\
\qquad\qquad "min\_star": \{"type": "integer", "description": "Minimum star rating (1-5)"\},\\
\qquad\qquad "amenity": \{"type": "string", "description": "Semantic search for amenities"\},\\
\qquad\qquad "sort\_by": \{"type": "string", "enum": ["price", "rating", "star"], "description": "Sort results by this field"\},\\
\qquad\qquad "top\_k": \{"type": "integer", "description": "Number of results per city (default 10)"\}\\
\qquad\quad \},\\
\qquad\quad "required": ["city"]\\
\qquad \}\\
\quad \},\\
\quad "search\_attractions": \{\\
\qquad "name": "search\_attractions",\\
\qquad "description": "Search for attractions in one or multiple cities. Use comma-separated cities for batch query.",\\
\qquad "parameters": \{\\
\qquad\quad "type": "object",\\
\qquad\quad "properties": \{\\
\qquad\qquad "city": \{"type": "string", "description": "City or comma-separated cities"\},\\
\qquad\qquad "attraction\_name": \{"type": "string", "description": "Look up a SPECIFIC attraction by name -- absence from a browse listing does NOT mean it is missing"\},\\
\qquad\qquad "max\_ticket\_price": \{"type": "number", "description": "Maximum ticket price"\},\\
\qquad\qquad "facility": \{"type": "string", "description": "Semantic search for facilities"\},\\
\qquad\qquad "sort\_by": \{"type": "string", "enum": ["price", "rating", "duration\_of\_visit"], "description": "Sort results by this field"\},\\
\qquad\qquad "top\_k": \{"type": "integer", "description": "Number of results per city (default 10)"\}\\
\qquad\quad \},\\
\qquad\quad "required": ["city"]\\
\qquad \}\\
\quad \},\\
\quad "search\_cars": \{\\
\qquad "name": "search\_cars",\\
\qquad "description": "Search for rental cars in one or multiple cities. Use comma-separated cities for batch query.",\\
\qquad "parameters": \{\\
\qquad\quad "type": "object",\\
\qquad\quad "properties": \{\\
\qquad\qquad "city": \{"type": "string", "description": "City or comma-separated cities"\},\\
\qquad\qquad "min\_capacity": \{"type": "integer", "description": "Minimum passenger capacity"\},\\
\qquad\qquad "max\_price\_per\_day": \{"type": "number", "description": "Maximum price per day"\},\\
\qquad\qquad "car\_type": \{"type": "string", "description": "Type of car"\},\\
\qquad\qquad "extra\_service": \{"type": "string", "description": "Semantic search for extra services"\},\\
\qquad\qquad "top\_k": \{"type": "integer", "description": "Number of results per city (default 10)"\}\\
\qquad\quad \},\\
\qquad\quad "required": ["city"]\\
\qquad \}\\
\quad \},\\
\quad "compute\_travel\_time": \{\\
\qquad "name": "compute\_travel\_time",\\
\qquad "description": "FREE (does NOT count against the 15-search budget). Distance + the MINIMUM door-to-door minutes the feasibility check (B3) requires between two locations -- the exact model the scorer uses. Identify each endpoint by \_latitude/\_longitude from a search result, or by \_name/\_city/\_type.",\\
\qquad "parameters": \{\\
\qquad\quad "type": "object",\\
\qquad\quad "properties": \{\\
\qquad\qquad "from\_name": \{"type": "string"\}, "from\_city": \{"type": "string"\},\\
\qquad\qquad "from\_type": \{"type": "string", "enum": ["attraction", "hotel", "flight"]\},\\
\qquad\qquad "from\_latitude": \{"type": "number"\}, "from\_longitude": \{"type": "number"\},\\
\qquad\qquad "to\_name": \{"type": "string"\}, "to\_city": \{"type": "string"\},\\
\qquad\qquad "to\_type": \{"type": "string", "enum": ["attraction", "hotel", "flight"]\},\\
\qquad\qquad "to\_latitude": \{"type": "number"\}, "to\_longitude": \{"type": "number"\}\\
\qquad\quad \},\\
\qquad\quad "required": []\\
\qquad \}\\
\quad \},\\
\quad "write\_note": \{\\
\qquad "name": "write\_note",\\
\qquad "description": "FREE (does NOT count against the search budget). Jot a short note to your planning notebook (a chosen flight+times, a day-to-city assignment, an opening-hours constraint). You will see all your notes again at final-planning time.",\\
\qquad "parameters": \{\\
\qquad\quad "type": "object",\\
\qquad\quad "properties": \{"text": \{"type": "string", "description": "The note to remember."\}\},\\
\qquad\quad "required": ["text"]\\
\qquad \}\\
\quad \},\\
\quad "submit\_plan": \{\\
\qquad "name": "submit\_plan",\\
\qquad "description": "Submit the final travel plan, or a refusal for a genuinely impossible task.",\\
\qquad "parameters": \{\\
\qquad\quad "type": "object",\\
\qquad\quad "properties": \{\\
\qquad\qquad "is\_feasible": \{"type": "boolean", "description": "Whether the travel plan is feasible"\},\\
\qquad\qquad "refusal\_reason": \{"type": "string", "description": "Reason for refusal if not feasible"\},\\
\qquad\qquad "plan": \{"type": "object", "description": "The travel plan organized by day"\}\\
\qquad\quad \},\\
\qquad\quad "required": ["is\_feasible", "plan"]\\
\qquad \}\\
\quad \}\\
\}

\#\# Complete system prompt (verbatim)

You are an expert travel-planning agent operating inside TREK, a self-contained travel SANDBOX.

\# The sandbox is ground truth --- trust it\\
- A tool-backed database of real-shaped flights, hotels, attractions, and rental cars IS the complete and authoritative world for this task. Every record the tools return is REAL and CORRECT within this world.\\
- NEVER refuse, hedge, or stop planning because the data ``looks synthetic / fake / wrong / incomplete''. Do not second-guess prices, names, times, or coordinates the tools give you. Your job is to PLAN with what the database contains, not to judge whether it matches the outside world.\\
- Use ONLY entities returned by the tools, with their EXACT names, times, and prices. Never invent a flight number, hotel, attraction, car, or a time/price that a tool did not give you.

\# Flights are DIRECT only\\
Every request specifies direct flights with no connections. An itinerary is a sequence of DIRECT flights between consecutive cities: departure city $\rightarrow$ city 1 $\rightarrow$ \ldots{} $\rightarrow$ city n ($\rightarrow$ departure city if it is a round trip). You may NOT satisfy a leg by routing through an intermediate city. If search\_flights returns nothing for a required leg, that leg cannot be flown.

\# When (and only when) to refuse\\
Set is\_feasible=false ONLY if the request is genuinely impossible in this sandbox, and say which of these it is:\\
\quad 1. budget --- no combination of real options fits the stated budget;\\
\quad 2. no route --- some required leg has no DIRECT flight (and connections are ruled out);\\
\quad 3. entity does not exist --- a specifically named place/hotel/etc. is not in the database.\\
Before claiming an entity does not exist, LOOK IT UP BY NAME (search\_hotels(name=\ldots) or search\_attractions(attraction\_name=\ldots)). A plain browse returns only the top\_k rows, so a name that is absent from a browse listing is NOT evidence that it is missing from the database.\\
State the cause explicitly in refusal\_reason --- say which of the three it is and name the entity, city, or leg involved. ``I cannot do this'' is not a diagnosis.\\
If you can build ANY valid plan that satisfies the hard constraints, you MUST submit it (is\_feasible=true). ``Hard to optimize'' is not a refusal.

\# Tools\\
- search\_flights / search\_hotels / search\_attractions / search\_cars --- gather options. You have 15 billable searches. A grounded plan needs about one search per city for each resource the task involves (one per flight leg, plus hotels / attractions / cars per city) --- searching too LITTLE is penalised exactly as heavily as searching too much, so cover every city and don't repeat identical searches. A round-trip flight search returns both legs and costs 2.\\
- compute\_travel\_time --- FREE. Returns the exact minimum door-to-door minutes the feasibility check uses between two places. Use it to confirm consecutive same-day events are reachable in time.\\
- write\_note --- FREE. Jot decisions to your notebook (chosen flights+times, day$\rightarrow$city plan, opening-hours limits). Your notes stay in this conversation --- re-read them here when you compose the plan.

\# Implicit needs --- the traveller description is a REQUIREMENT, not flavour text\\
Every request describes who is travelling (``for luxury travelers'', ``for foodies'', ``with pets'', ``with children'', ``elderly travelers'', ``business travelers'', ``solo women'', ``photography'', ``nightlife enthusiast'', ``road trip'', ``couples trip'', ``disabled traveler'', ``fast-paced budget travel''). Each of these implies concrete needs that your bookings must actually satisfy, and your plan is scored on whether they do.\\
Two of these carry a concrete BOOKING implication, not just a preference:\\
- ``road trip'' means the traveller drives: book a rental car in every stay city, for every day of the trip, even though the request does not spell out a car.\\
- Any traveller type that names a mobility or accessibility need must be met by the resource you book, not by the itinerary alone.\\
Work out for yourself which amenities / facilities / services each traveller type needs, then pick resources that CARRY them: the search tools let you filter by \texttt{amenity} (hotels), \texttt{facility} (attractions) and \texttt{extra\_service} (cars), and every result lists its amenities/facilities in full, so check them before booking. A hotel that ignores the stated traveller type is a worse answer than one that matches it, even if both fit the budget.

\# Feasibility rules the scorer enforces (plan for them)\\
- Opening hours: an attraction visit [visit\_start, visit\_end] must fall within its open\_hours, and visit\_end $-$ visit\_start should match its duration\_of\_visit.\\
- Spatio-temporal: between two consecutive same-day events, the time gap must be $\geq$ compute\_travel\_time's minimum for that hop. Flights: schedule around their departure\_time/arrival\_time; don't place an event before you've physically arrived.\\
- Budget \& party size: total cost across flights + hotels (per night, enough rooms) + attractions + car (per day) must respect the budget and passenger count.\\
- Sightseeing coverage: every city you SLEEP in needs at least one scheduled attraction visit, so schedule at least as many visits as there are stay cities.\\
- Time format: every time you write --- departure\_time, arrival\_time, visit\_start, visit\_end, and the hotel's check\_in --- must be a 24-hour clock time \texttt{HH:MM}. The day is already given by the day key; ``day1'' is not a check-in time.

\# Workflow\\
1. Read the request; note the cities, days, budget, party size, and any special requirement.\\
2. Search each required city for the resources the request needs (flights, hotels, attractions, cars). Record good candidates with write\_note.\\
3. Use compute\_travel\_time to sanity-check the day-by-day timing.\\
4. When you have enough, compose the FINAL day-by-day plan yourself and call submit\_plan. Nothing will prompt you to start --- decide for yourself when you have searched enough.

Make every decision yourself. Never ask the user anything.

\# Plan structure passed to submit\_plan\\
Day-keyed object (\texttt{day1}, \texttt{day2}, \ldots). Each day: current\_city (``A'' or ``A to B''); flights [\{flight\_number, departure\_city, arrival\_city, departure\_time, arrival\_time, price\}]; attractions [\{name, city, visit\_start, visit\_end\}]; hotel \{name, city, price\_per\_night, check\_in\}; car \{car\_type, capacity, price\_per\_day, city\}. Use ONLY entities returned by the search tools, with their exact names and times; every time field is \texttt{HH:MM}. A refusal is submit\_plan(is\_feasible=false, refusal\_reason=\ldots, plan=\{\}).
\end{scriptsize}
\end{tcolorbox}

\subsection{Implicit Need Understanding}
\label{ssec:implicit-need-understanding}

We analyze the most frequently detected implicit demand keywords across all models under our unified Bedrock-native function-calling agent. The most commonly surfaced keywords are ``luxury travelers,'' ``road trip,'' and ``disabled traveler,'' indicating that models are particularly sensitive to high-end, driving, and accessibility needs. These patterns suggest that tool-using agents can infer diverse unstated preferences from persona descriptions, though coverage varies by model---stronger models detect a broader range of implicit needs while weaker models concentrate on a narrower subset of high-salience keywords.

\subsection{Per-Dimension Failure Rates}
\label{app:failure-rates}
\begin{figure*}[t]
  \centering
  \includegraphics[width=0.72\textwidth]{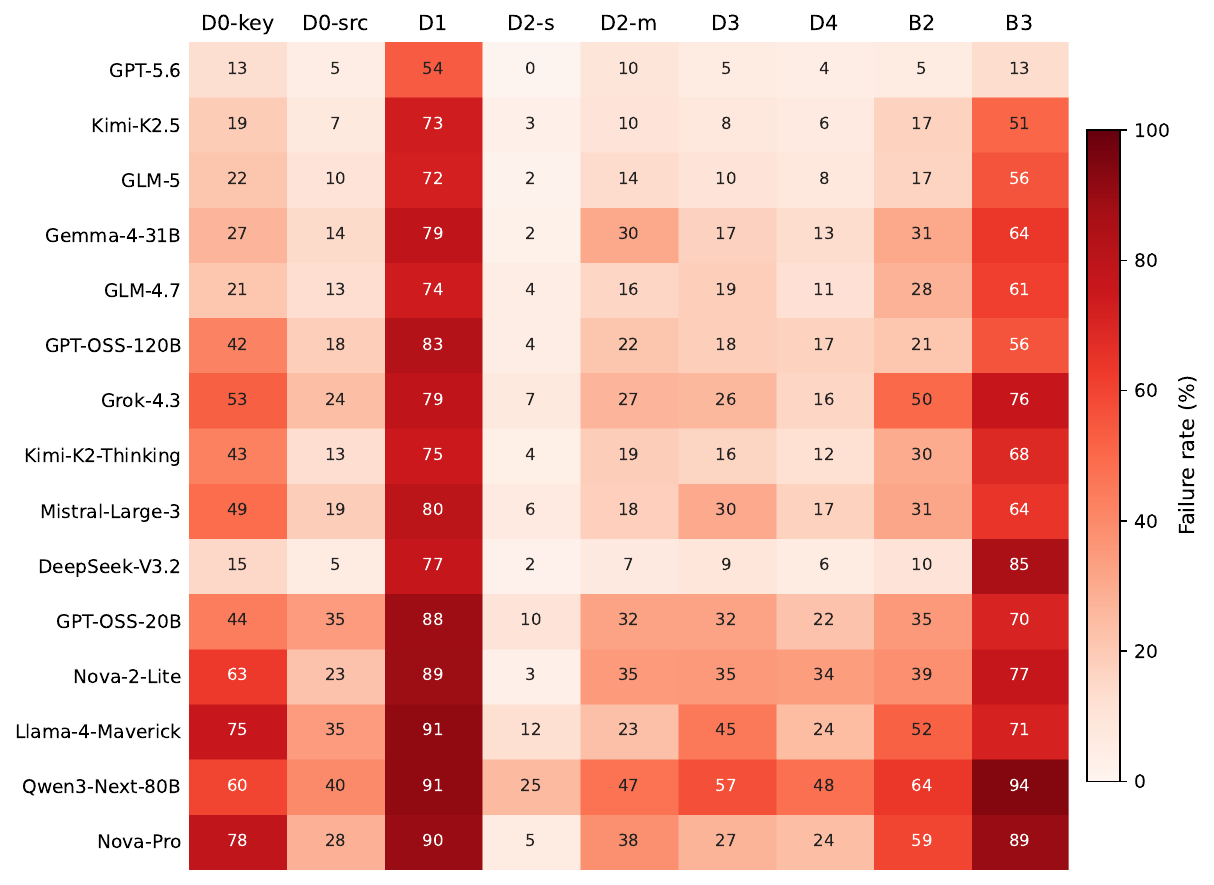}
  \caption{Failure decomposition: per-dimension failure rate (\% of applicable tasks scoring $<1.0$), $15$ agents (rows, sorted by TP-feas) $\times$ $9$ correctness dimensions. \textbf{D1} (implicit needs, boxed) is a top-two failure for \emph{every} agent and the sole dimension the frontier model still fails at scale ($54\%$ vs.\ its next-worst $13\%$). \textbf{B3} (spatio-temporal reachability, boxed) broadly darkens down the ranking---the planner/non-planner watershed. The bottleneck rises with capability: weak agents fail everywhere, the frontier fails essentially only D1.}
  \label{fig:failure_heatmap}
\end{figure*}

Figure~\ref{fig:failure_heatmap} visualizes the per-dimension failure rate for every model as a heatmap, and Table~\ref{tab:failure_decomp} lists the exact underlying numbers.

% Failure decomposition — per-dimension FAIL rate (%). Generated from analysis/leaderboard_v4.csv (fail_ recomputed from trek_scores per-task). Do not hand-edit.
\begin{table*}[t]
  \centering \small \setlength{\tabcolsep}{4.5pt}
  \caption{Failure decomposition: per-dimension failure rate (\%, fraction of applicable tasks scoring $<1.0$ on that dimension), sorted by TP-feas; \textbf{bold} marks a cell whose failure rate exceeds $50\%$. These are per-dimension \emph{failure} rates; the four categories in Table~\ref{tab:main_results} are instead \emph{all-or-nothing} (a task counts only if it passes every dimension in the group), with D1 folded into the Satisfaction category there rather than shown as its own column. \textbf{D1 (implicit needs) is a top-two failure for every one of the 15 agents} and the sole dimension the frontier model still fails at scale; \textbf{B3 (spatio-temporal reachability)} is the planner/non-planner watershed (13.3\% at the top to 93.8\% for the weakest). The \textbf{D4} column is the feasibility-classification failure over all $800$ tasks; the Inf category in Table~\ref{tab:main_results} instead scopes D4 to the $267$ infeasible tasks. Rows are the same 15 models as Table~\ref{tab:main_results}, three abbreviated for width (Mistral-L3 = Mistral-Large-3, Kimi-K2-Think = Kimi-K2-Thinking, Llama-4-Mav = Llama-4-Maverick).}
  \label{tab:failure_decomp}
  \begin{tabular*}{\textwidth}{@{\extracolsep{\fill}} l ccccccccc @{}}
    \toprule
    \textbf{Model} & D0-key & D0-src & D1 & D2-s & D2-m & D3 & D4 & B2 & B3 \\
    \midrule
    GPT-5.6 & 12.8 & 5.1 & \textbf{53.7} & 0.4 & 9.6 & 5.3 & 4.4 & 5.4 & 13.3 \\
    Kimi-K2.5 & 19.3 & 7.1 & \textbf{73.1} & 3.2 & 9.9 & 8.3 & 6.5 & 16.9 & \textbf{50.7} \\
    GLM-5 & 21.8 & 9.9 & \textbf{72.2} & 1.6 & 13.8 & 10.1 & 8.2 & 16.5 & \textbf{55.7} \\
    Gemma-4-31B & 26.8 & 14.3 & \textbf{79.3} & 2.4 & 30.5 & 17.3 & 13.4 & 30.6 & \textbf{63.8} \\
    GLM-4.7 & 21.2 & 12.9 & \textbf{73.6} & 4.4 & 16.0 & 18.6 & 11.4 & 27.6 & \textbf{61.0} \\
    GPT-OSS-120B & 42.2 & 18.4 & \textbf{82.6} & 4.4 & 21.6 & 18.4 & 17.0 & 21.4 & \textbf{55.5} \\
    Grok-4.3 & \textbf{52.5} & 23.8 & \textbf{78.9} & 7.2 & 26.6 & 25.9 & 15.9 & \textbf{50.3} & \textbf{76.2} \\
    Kimi-K2-Think & 42.6 & 13.1 & \textbf{74.9} & 3.6 & 18.8 & 16.3 & 11.9 & 30.0 & \textbf{68.5} \\
    Mistral-L3 & 49.0 & 18.8 & \textbf{79.7} & 6.4 & 18.1 & 30.2 & 17.0 & 31.3 & \textbf{64.2} \\
    DeepSeek-V3.2 & 15.0 & 5.4 & \textbf{77.3} & 2.0 & 6.7 & 8.6 & 5.9 & 10.1 & \textbf{85.4} \\
    GPT-OSS-20B & 43.9 & 34.5 & \textbf{88.3} & 10.0 & 32.3 & 32.3 & 22.5 & 35.5 & \textbf{70.4} \\
    Nova-2-Lite & \textbf{63.2} & 22.5 & \textbf{89.0} & 3.2 & 35.1 & 35.5 & 34.0 & 38.6 & \textbf{76.7} \\
    Llama-4-Mav & \textbf{75.4} & 34.7 & \textbf{91.2} & 12.4 & 23.4 & 45.0 & 24.5 & \textbf{52.2} & \textbf{71.3} \\
    Qwen3-Next-80B & \textbf{59.7} & 40.0 & \textbf{91.2} & 25.1 & 47.2 & \textbf{56.7} & 48.0 & \textbf{64.0} & \textbf{93.8} \\
    Nova-Pro & \textbf{78.2} & 27.8 & \textbf{89.6} & 5.2 & 37.9 & 27.0 & 23.8 & \textbf{59.5} & \textbf{89.5} \\
    \bottomrule
  \end{tabular*}
\end{table*}

\subsection{Failure Mode Taxonomy}
\label{app:failure-taxonomy}
The per-dimension rates in Table~\ref{tab:failure_decomp} correspond to five qualitatively distinct, recurring failure modes; we describe each with its scored dimension and a representative case observed in our runs.

\begin{itemize}[leftmargin=*, itemsep=2pt]
  \item \textbf{Entity hallucination (D0-src).} The agent books a plausible-sounding entity absent from the sandbox KB, typically by falling back on real-world knowledge when the search results do not match its expectations. \emph{Example} (Nova-Pro, a trip to Istanbul and Thessaloniki): the agent correctly books the KB-listed hotels, then schedules ``Hagia Sophia'' and the ``White Tower of Thessaloniki''---famous \emph{real} landmarks that do not exist in the synthetic KB---which zeroes D0-src. This is exactly the mode the sandbox framing (\S\ref{app:system-prompt}) is designed to suppress.
  \item \textbf{Implicit-need miss (D1).} The booked resources meet the explicit request but not the persona's \emph{unstated} needs (e.g., no pet-friendly hotel for a ``with pets'' traveler). This is the universal bottleneck (\S\ref{ssec:rq2}) and dominates even the frontier model.
  \item \textbf{Multi-city omission (D2-multi).} A required destination is dropped: the agent produces a coherent plan through a \emph{subset} of the requested cities.
  \item \textbf{Budget overrun (D3).} The total booked cost exceeds the stated budget, scored by the exponential-decay penalty of \S\ref{ssec:scoring-dimensions}.
  \item \textbf{Infeasible transition (B3).} Consecutive same-day events are scheduled with less time between them than the minimum door-to-door travel time, yielding a day no traveler could physically execute.
\end{itemize}
\noindent Table~\ref{tab:failure_decomp} gives the per-model prevalence of each mode: weak models exhibit all five, whereas the strongest agent has driven every mode down except the implicit-need miss.

\begin{figure}[t]
  \centering
  \includegraphics[width=0.86\columnwidth]{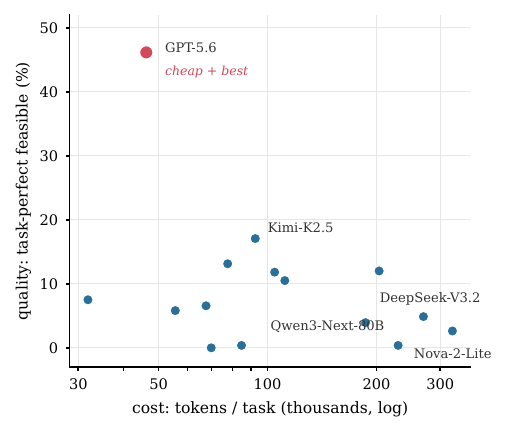}
  \caption{Cost vs.\ quality across the $15$ agents: per-task token cost (log scale) against task-perfect-feasible rate. There is no cost--quality correlation---GPT-5.6 is both the best and the cheapest capable agent, while the weakest agents are among the most expensive.}
  \label{fig:cost_quality}
\end{figure}

\subsection{Leaderboard Robustness}
\label{app:robustness}
The headline \emph{task-perfect} rate is a per-task binary aggregated as a proportion, so its sampling uncertainty is a binomial standard error over the $533$ feasible tasks; Table~\ref{tab:robustness} reports each model's TP-feas with a $95\%$ normal-approximation interval. The four correctness categories are combined by a geometric mean (\S\ref{ssec:composite-scores}); to show the leaderboard does not hinge on that choice, the table also gives the ``Overall'' composite under arithmetic and harmonic means. The induced ranking is nearly invariant to the aggregation: Spearman's $\rho$ against the geometric-mean order is $0.96$ (arithmetic), $0.996$ (harmonic), and $0.97$ (worst-category minimum), and $0.97$ against the TP-feas order; GPT-5.6 is first under every aggregation and the top three are unchanged. The one notable move is DeepSeek-V3.2 (arithmetic rank $5\!\rightarrow\!$ geometric rank $8$): its high truthfulness and refusal scores mask a near-zero executability, which the geometric mean---by design---does not let one strong category hide. The TP-feas intervals confirm the headline gaps far exceed sampling noise: the $46.2\%$ frontier is separated from the $17.1\%$ runner-up by many standard errors, and only adjacent near-ties (e.g., Llama-4-Maverick and Qwen3-Next-80B, both $0.4\%$) overlap---exactly the ``read one-to-two-task gaps as ties'' caveat of Appendix~\ref{app:limitations}.

\begin{table}[t]
  \centering \small \setlength{\tabcolsep}{3.5pt}
  \caption{Leaderboard robustness. TP-feas (\%) with a $95\%$ interval, and the \emph{Overall} composite of the four correctness categories under arithmetic / geometric / harmonic means (\%). Rows sorted by TP-feas; the geometric mean is the one used in \S\ref{ssec:composite-scores}. The ranking is near-invariant to the aggregation choice (Spearman $\rho\geq0.96$).}
  \label{tab:robustness}
  \begin{tabular}{lccccc}
    \toprule
    \textbf{Model} & \textbf{TP-feas} & \textbf{95\% CI} & \textbf{Ovr$_\text{arith}$} & \textbf{Ovr$_\text{geo}$} & \textbf{Ovr$_\text{harm}$} \\
    \midrule
    GPT-5.6          & 46.2 & [41.9,\,50.4] & 82.2 & 79.7 & 76.7 \\
    Kimi-K2.5        & 17.1 & [13.9,\,20.3] & 66.0 & 59.8 & 53.9 \\
    GLM-5            & 13.1 & [10.3,\,16.0] & 63.5 & 57.1 & 51.2 \\
    Gemma-4-31B      & 12.0 & [\,9.2,\,14.8] & 56.1 & 48.0 & 41.0 \\
    GLM-4.7          & 11.8 & [\,9.1,\,14.6] & 59.4 & 52.7 & 46.7 \\
    GPT-OSS-120B     & 10.5 & [\,7.9,\,13.1] & 54.3 & 47.1 & 39.7 \\
    Grok-4.3         & \,7.5 & [\,5.3,\,\,9.7] & 50.5 & 39.4 & 30.4 \\
    Kimi-K2-Thinking & \,6.6 & [\,4.5,\,\,8.7] & 55.8 & 46.2 & 38.2 \\
    Mistral-Large-3  & \,5.8 & [\,3.8,\,\,7.8] & 50.1 & 40.9 & 32.5 \\
    DeepSeek-V3.2    & \,4.9 & [\,3.0,\,\,6.7] & 57.5 & 43.7 & 31.8 \\
    GPT-OSS-20B      & \,3.9 & [\,2.3,\,\,5.6] & 44.5 & 35.2 & 26.6 \\
    Nova-2-Lite      & \,2.6 & [\,1.3,\,\,4.0] & 33.2 & 24.2 & 18.1 \\
    Llama-4-Maverick & \,0.4 & [\,0.0,\,\,0.9] & 38.3 & 23.6 & 10.8 \\
    Qwen3-Next-80B   & \,0.4 & [\,0.0,\,\,0.9] & 23.5 & 14.0 & \,8.7 \\
    Nova-Pro         & \,0.0 & [\,0.0,\,\,0.0] & 36.0 & 19.6 & 10.6 \\
    \bottomrule
  \end{tabular}
\end{table}

\section{Source Reference}
\label{app:source-reference}
This section details the origins of all datasets used, explains our rationale for selecting implicit demand keywords within the data, and describes how those implicit demands guided our choice of hotels and attractions.

\subsection{Data layer}
\label{app:data-layer}
\subsubsection{Airports data (\texttt{airports.csv})}
\label{sec:source-airports}

\begin{itemize}[leftmargin=*]
  \item \textbf{File}: \texttt{airports.csv}
  \item \textbf{Source}: OurAirports, \url{https://ourairports.com/data/}
  \item \textbf{Contents}: Global airport metadata, including IATA/ICAO codes, names, geographic coordinates, airport type, and operational status.
  \item \textbf{Purpose}:  
    To identify all \textit{type} entries with \textit{scheduled\_service} status that have at least one scheduled flight connection to another large airport.
  \item \textbf{Processing steps}:  
    \begin{enumerate}[label=(\alph*)]
      \item Filter rows where type = ``large\_airport'', scheduled\_service = ``yes.''

      \item Cross-reference each remaining airport’s IATA/ICAO code against our flight-segments dataset to ensure it connects to at least one other selected large airport.
      \item Map each airport to its corresponding city using its nearest major city field or geographic coordinates.
      \item Use the resulting city list to seed the creation of our attractions, hotels, and flight-segments CSV files.
    \end{enumerate}
\end{itemize}

\subsubsection{Implicit Demand Keywords}
\label{sec:implicit-keywords}

This subsection explains, for each of the 13 implicit demand keywords, the underlying traveler need it captures, the evidence for treating it as a real segment, and how tagging it in our system yields concrete benefits (e.g., more relevant recommendations, higher user satisfaction).

\paragraph{What the cited evidence does and does not establish.}
The sources below establish that each segment exists and that its broad concern is documented---families weigh children's facilities, travelers with mobility disabilities meet accessibility barriers, food experiences shape destination impressions. They do \emph{not} establish the specific facility list we attach to each persona: no survey reports the share of travelers who require, say, a pet relief area or a roll-in shower in particular. Those lists are our \emph{operationalisation}, chosen so that each requirement is a fixed, published, machine-checkable field in the knowledge base. What matters for the benchmark is not that the mapping is empirically derived but that it is deterministic and auditable: it is identical for every agent, released with the evaluator, and open to being contested and revised. Where the evidence is thin or the sampling is weak we say so in-line rather than smoothing it over.

\begin{description}[leftmargin=0pt,labelwidth=3cm,style=nextline]

  \item[With children]  
  Families traveling with young children overwhelmingly prioritize accommodations and attractions that cater to kids’ needs. “70\% of families consider children’s amenities---baby facilities, play areas, supervised entertainment---an important factor when choosing a destination” \citep{condorferries_family2025}.  
  \textbf{Rationale:} By tagging “with children,” our system filters out venues lacking playgrounds, cribs, or children’s programs.  
  \textbf{Benefit:} Ensures families see only truly kid-friendly options, reducing browsing time and boosting booking conversion among family segments.

  \item[Road trip]  
  Self-driving tourists face parking and vehicle-access challenges: in a 2020 CheLun-platform survey reported in a joint Tuniu--CheLun report, $70.43\%$ of surveyed users identified scarce parking as a troublesome problem encountered when parking on self-drive trips~\citep{chinadaily_2020}.  
  \textbf{Rationale:} Tagging “road trip” prioritizes properties with guaranteed on-site or nearby secure parking, easy drive-in access, and luggage drop-off services.  
  \textbf{Benefit:} Dramatically reduces traveler frustration, leading to higher ratings and repeat bookings for driving itineraries.

  \item[Elderly travelers]  
  Mobility limitation rises sharply with age: $26.9\%$ of U.S.\ adults aged $65+$ have a mobility disability---serious difficulty walking or climbing stairs---against $13.7\%$ of adults overall~\citep{okoro2018disability}. Barrier-free features are therefore disproportionately relevant to older travelers, which is what this tag encodes; we do \emph{not} treat age as equivalent to disability, and the ``disabled traveler'' tag below is scored separately.  
  \textbf{Rationale:} The “elderly travelers” tag surfaces hotels with ground-floor rooms, roll-in showers, elevators, and attractions with wheelchair ramps.  
  \textbf{Benefit:} Addresses critical accessibility needs, improving comfort and safety metrics and expanding inclusivity.

  \item[Business travelers]  
  Hotel guests rank fast, reliable Wi-Fi as the single most important amenity ($94\%$ in a 2013 Forrester survey), and $34\%$ of business travelers call its absence a deal-breaker~\citep{comcast_2014}.  
  \textbf{Rationale:} Tagging “business travelers” elevates listings offering 24/7 business centers, high-speed internet, meeting rooms, and express check-in/out.  
  \textbf{Benefit:} Meets the top priorities of corporate guests, shortening decision time and increasing satisfaction scores.

  \item[With pets]  
  Pet owners treat animals as family, and pet travel is still growing: among dog owners who travel with their dogs, $74\%$ flew with them in the past year---up from $68\%$ in 2023---and $87\%$ travelled with them by car~\citep{appa_2025}; $53\%$ of travelers report taking holidays with pets~\citep{condorferries_pets2025}.  
  \textbf{Rationale:} The “with pets” filter highlights accommodations with pet relief areas, pet-welcome packages, and nearby dog parks.  
  \textbf{Benefit:} Captures a fast-growing niche, increasing bookings and positive reviews from pet-owning travelers.

  \item[Nightlife enthusiast]  
  Evening entertainment is a studied driver of urban attractiveness for younger segments: \citet{lee2024nightlife} evaluate nightlife attractiveness across $161$ Seoul districts specifically for Millennials and Generation Z. We therefore treat nightlife as a persona axis, without claiming a population share for it.  
  \textbf{Rationale:} Tagging “nightlife enthusiast” surfaces hotels near entertainment districts, late-night dining, and attractions with evening events.  
  \textbf{Benefit:} Boosts engagement and length of stay among Millennials and Gen Z by delivering curated after-dark experiences.

  \item[Disabled traveler]  
  Travelers with disabilities face widespread barriers: 96\% report accommodation issues, 81\% inaccessible bathrooms, and 84\% want pre-trip accessibility information \citep{mmgy_2022}.  
  \textbf{Rationale:} The “disabled traveler” tag promotes ADA-compliant rooms, tactile signage, visual alarms, and attractions with guided-access options.  
  \textbf{Benefit:} Reduces booking friction and enhances satisfaction for a sizable and often underserved traveler segment.

  \item[Fast-paced budget travel]  
  Ultra-efficient, low-cost itineraries---``special forces travel''---appeal strongly to young adventurers: in a voluntary online poll of roughly $800$ respondents, $82.18\%$ were aged $20$--$22$, and $75.94\%$ said the style appealed to them because it is time-efficient (a further $12.61\%$ cited cost-effectiveness)~\citep{yicai_2023}.  
  \textbf{Rationale:} Tagging “fast-paced budget travel” prioritizes hostels or capsule hotels with express check-in, central locations, and multi-attraction passes.  
  \textbf{Benefit:} Enables high-velocity exploration at minimal cost, increasing appeal among students and backpackers.

  \item[Couples trip]  
  Romantic getaways emphasize intimacy and pampering: in an April 2022 OnePoll survey of $1{,}701$ U.S.\ adults who were married, cohabiting or in a relationship, the most popular honeymoon itinerary included romantic activities such as hot-air-balloon rides or sunset cruises ($54\%$) and pampering ($52\%$)~\citep{expedia_2022}.  
  \textbf{Rationale:} The “couples trip” filter surfaces boutique hotels with honeymoon packages, private dining, and couples’ spa services.  
  \textbf{Benefit:} Elevates guest experience and average booking spend through targeted romantic offerings.

  \item[Solo women]  
  Safety concerns drive travel choices: $76\%$ of women travellers say they would feel unsafe holidaying by themselves, and $75\%$ of women who have never travelled solo would be more likely to do so on a group tour~\citep{condorferries_solo2025}. (These figures come from an industry compilation that does not publish its sampling or field dates; they motivate the segment, not the specific amenity list below.)  
  \textbf{Rationale:} Tagging “solo women” highlights women-only floors, group-friendly excursions, and 24/7 staffed properties.  
  \textbf{Benefit:} Boosts confidence and uptake by addressing safety and social comfort needs.

  \item[Luxury travelers]  
  High-end travelers are characterised less by spend than by expectations of exclusivity, privacy and personalised service, and two of the signals they rely on map directly onto bookable fields. In McKinsey's 2024 State of Tourism Survey ($5{,}061$ travelers across five markets, fielded 27 February--11 March 2024; luxury defined as ${\geq}\$500$ per night on lodging), $84\%$ of luxury travelers said government-assigned \emph{star rankings} matter when choosing accommodation, against $66\%$ of mass travelers, and $77\%$ versus $53\%$ said the same of hotel brand; $88\%$ of respondents rated \emph{fitness} important on leisure trips~\citep{mckinsey_2024}. The same source cautions that luxury wellness expectations now extend well beyond a spa, which is why our facility set treats spa, gym, premium bedding and bar as alternatives rather than requiring any one of them.  
  \textbf{Rationale:} The “luxury travelers” tag elevates recommendations to include butler service, private transfers, and Michelin-starred experiences.  
  \textbf{Benefit:} Maximizes satisfaction and revenue per booking through premium, bespoke offerings.

  \item[Foodie]  
  Culinary experiences drive trip decisions: 81\% of travelers most look forward to sampling local cuisines, and 83\% believe food experiences shape destination perception \citep{hotelbeds_2024,foodinspiration_2023}.  
  \textbf{Rationale:} Tagging “foodie” highlights food tours, street-food markets, cooking classes, and hotels with renowned restaurants.  
  \textbf{Benefit:} Enhances itinerary engagement by catering to gastronomy-driven travelers.

  \item[Photography]  
  Photo-taking measurably shapes the travel experience: \citet{lee2021phototaking} find across five studies that taking photographs raises satisfaction with an experience while \emph{lowering} intention to revisit it, and that venue photography policy affects both. The tag therefore encodes photo-friendliness of venues, not a claim that photogenicity drives destination choice.  
  \textbf{Rationale:} The “photography” tag surfaces scenic viewpoints, photo-friendly museums, and hotels with rooftop terraces.  
  \textbf{Benefit:} Maximizes social-media shareability and initial booking interest among photography enthusiasts.

\end{description}
\subsubsection{Facilities and Services for Implicit Demand Keywords}
\label{sec:keyword-facilities}

Based on the implicit demand keywords defined above, we have identified the corresponding attractions facilities, hotel amenities, and vehicle services that best satisfy each need. The following details, compiled from industry reports and service providers, will be summarized in Appendix~\ref{app:facilities}.

\begin{description}[leftmargin=0pt,labelwidth=3cm,style=nextline]

  \item[With children]  
  \begin{itemize}[leftmargin=*]
    \item \textbf{Hotel Amenities}: Baby cots, roll-away beds for older children, infant bathtubs, outlet covers; children’s TV channels; stroller-accessible corridors, stroller storage areas, spacious elevators; indoor playgrounds or game rooms; welcome gifts for kids \citep{pmc9325268}.
    \item \textbf{Attraction Facilities}: Parent–child restrooms or family toilets; stroller-friendly ramps and pathways; on-site nursing rooms with bottle-warming stations and changing tables \citep{pmc9325268}.
    \item \textbf{Vehicle Services}: Child-safety seats (infant, toddler, booster) available as add-ons from major rental firms (Hertz, Avis, Alamo) meeting safety standards upon request \citep{avis_childseat,hertz_childseat}.
  \end{itemize}

  \item[Road trip]  
  \begin{itemize}[leftmargin=*]
    \item \textbf{Hotel Amenities}: Ample complimentary parking at motels or highway-side hotels; clear signage to parking areas \citep{travelersunited_parking}.
    \item \textbf{Attraction Facilities}: Large on-site parking lots; designated scenic pull-over bays with viewpoints; clear self-drive route wayfinding \citep{hoteliga_roadtrip}.
    \item \textbf{Vehicle Services}: Roadside assistance programs (mechanical breakdown, flat tires, lock-outs) included or optional through rental agencies \citep{hertz_roadside}.
  \end{itemize}

  \item[Elderly travelers]  
  \begin{itemize}[leftmargin=*]
    \item \textbf{Hotel Amenities}: Step-free entrances, elevators; non-slip flooring; grab bars in bathrooms; roll-in showers or seat-style showers \citep{texaslodging_senior}.
    \item \textbf{Attraction Facilities}: Ample seating and rest areas along paths; smooth, gently graded walkways; shaded benches \citep{texaslodging_senior}.
    \item \textbf{Vehicle Services}: Vehicles equipped with ADAS features—lane departure warning, blind-spot monitoring, automatic emergency braking—to reduce driving stress \citep{aaafoundation_adas}.
  \end{itemize}

  \item[Business travelers]  
  \begin{itemize}[leftmargin=*]
    \item \textbf{Hotel Amenities}: High-speed free Wi-Fi; spacious in-room workstations with desks and multiple outlets; refrigerators; 24/7 business center offering printing, copying, courier services; meeting and board rooms \citep{travelersunited_wifi,gourmetmarketing_business}.
    \item \textbf{Attraction Facilities}: Business-friendly lounges with Wi-Fi access and power outlets; quick-service cafés near meeting venues.
    \item \textbf{Vehicle Services}: GPS navigation with live traffic; airport express drop-off/pick-up; mobile Wi-Fi hotspots available for rent.
  \end{itemize}

  \item[With pets]  
  \begin{itemize}[leftmargin=*]
    \item \textbf{Hotel Amenities}: Pet beds and bowls; pet mats; welcome treats (dog biscuits, cat toys); curated pet-room service menus; information on nearby veterinary clinics and pet stores \citep{wander_petfriendly}.
    \item \textbf{Attraction Facilities}: Lists of nearby dog-friendly parks, trails, cafés with outdoor pet seating \citep{wander_petfriendly}.
    \item \textbf{Vehicle Services}: Pet-friendly rental policies (crate requirement, no-hair cleaning); pet safety harnesses or seat-belt attachments; scheduled stops for pet relief and water \citep{enterprise_pet}.
  \end{itemize}

  \item[Nightlife enthusiast]  
  \begin{itemize}[leftmargin=*]
    \item \textbf{Hotel Amenities}: 24-hour front desk; late-night room service; on-site bar, rooftop lounge, or nightclub \citep{siteminder_nightlife}.
    \item \textbf{Attraction Facilities}: Night markets, food streets, bars, clubs, midnight shows; extended evening hours \citep{cyberpublicity_nightlife}.
    \item \textbf{Vehicle Services}: Night-time public transit or “Night Tube” metro services on weekends/Fridays–Saturdays to support late-night guests \citep{tfl_nighttube}.
  \end{itemize}

  \item[Disabled traveler]  
  \begin{itemize}[leftmargin=*]
    \item \textbf{Hotel Amenities}: Wheelchair-accessible ramps or lifts at entrances; wide doorways; grab rails; accessible elevators; Braille signage; visual fire alarms \citep{numberanalytics_accessible}.
    \item \textbf{Attraction Facilities}: Ramps, wheelchair lifts, curb-cut paths; accessible restrooms; guided-access programs \citep{numberanalytics_accessible}.
    \item \textbf{Vehicle Services}: Fixed wheelchair tie-downs; dedicated disabled seats; audio-visual wayfinding in stations and vehicles.
  \end{itemize}

  \item[Fast-paced budget travel]  
  \begin{itemize}[leftmargin=*]
    \item \textbf{Hotel Amenities}: Budget hostels, capsule hotels, dormitory beds; shared bathrooms; free Wi-Fi for on-the-go planning \citep{gourmetmarketing_backpackers}.
    \item \textbf{Attraction Facilities}: Free or value-priced attractions (city walking tours, free-museum days); luggage storage or lockers; city passes covering multiple sites \citep{gourmetmarketing_backpackers}.
    \item \textbf{Vehicle Services}: No special requirements beyond basic rental.
  \end{itemize}

  \item[Couples trip]  
  \begin{itemize}[leftmargin=*]
    \item \textbf{Hotel Amenities}: King-size beds; in-room tubs or Jacuzzis; private balconies or villas with views; romantic packages (rose petals, champagne, chocolates); dual-treatment spa suites \citep{lovu_couples}.
    \item \textbf{Attraction Facilities}: Scenic viewpoints, beachfronts, garden parks ideal for couples; curated romantic photo spots.
    \item \textbf{Vehicle Services}: No special requirements.
  \end{itemize}

  \item[Solo women]  
  \begin{itemize}[leftmargin=*]
    \item \textbf{Hotel Amenities}: Women-only floors with controlled access; female staff on dedicated floors; in-room safes; well-lit corridors \citep{covington_solofloors}.
    \item \textbf{Attraction Facilities}: Popular, well-secured sites with guided small-group tours; avoidance of isolated areas.
    \item \textbf{Vehicle Services}: Designated women’s waiting areas at airports/stations; visible security presence.
  \end{itemize}

  \item[Luxury travelers]  
  \begin{itemize}[leftmargin=*]
    \item \textbf{Hotel Amenities}: Five-star resorts; spacious suites with premium bedding; high-end bath products; spa/Sauna; state-of-the-art fitness centers; infinity pools \citep{sawgrass_luxury}.
    \item \textbf{Attraction Facilities}: Private after-hours museum tours; VIP safari lodges; access to exclusive clubs; personalized cultural experiences (private cooking classes, wine tastings).
    \item \textbf{Vehicle Services}: Chauffeured luxury vehicles (Mercedes, Rolls-Royce) for airport transfers and excursions.
  \end{itemize}

  \item[Foodie]  
  \begin{itemize}[leftmargin=*]
    \item \textbf{Hotel Amenities}: On-site signature restaurants, especially Michelin-starred; guest chef events; in-room gourmet dining experiences \citep{fourseasons_gourmet}.
    \item \textbf{Attraction Facilities}: Local food markets, street-food tours, winery/brewery visits, cooking classes; guaranteed access to tasting events \citep{aaa_foodie}.
    \item \textbf{Vehicle Services}: No special requirements.
  \end{itemize}

  \item[Photography]  
  \begin{itemize}[leftmargin=*]
    \item \textbf{Hotel Amenities}: High-floor rooms with panoramic views; rooftop terraces; “Instagram Butler” service to assist with photography setups \citep{herie_instagrambutler}.
    \item \textbf{Attraction Facilities}: Designated photo platforms; timed-entry slots for clear shots; permits for equipment and drone use.
    \item \textbf{Vehicle Services}: Remote-shoot stops on scenic routes; vehicle-mounted camera mounts available on request.
  \end{itemize}

\subsubsection{Restaurant Rating Trends by Attraction Type}
\label{sec:restaurant-trends}

The table below summarizes our analysis of restaurant ratings around different attraction categories.  Inline citations use Author–Year format; full references appear at the end of this section.

\paragraph{Museum / Art Gallery}  
Restaurants near museums and galleries tend to earn higher ratings, typically clustering around 4.0–4.5 stars.  Museums are often located in well-developed urban cores with mature culinary scenes, and many maintain on-site cafés with strong reputations.  For example, the Museum Café at Toledo Museum of Art scores 4.5/5 on TripAdvisor, and the Museum Café at the Art Institute of Chicago averages about 4.0/5 \citep{TripAdvisor_Toledo2025,TripAdvisor_Chicago2025}.  Visitors to cultural institutions also skew toward leisure-oriented, quality-seeking diners, resulting in lower negative review rates and consistently high average scores.

\paragraph{Historic Sites / Landmarks}  
Ratings near historic landmarks exhibit a bimodal distribution due to the “tourist-trap” phenomenon.  High foot traffic sustains many average-quality eateries (often rated 3.0–4.0), while a few premium establishments approach 4.5–5.0 stars.  Investigations have shown that inexperienced tourists frequently overrate mediocre venues, skewing ratings upward \citep{Reddit_TouristTrap2013}.  Overall, mean ratings around historic sites settle between 3.5 and 4.0, with caution advised regarding review authenticity.

\paragraph{Nature / Park}  
Dining options within parks and natural attractions generally yield moderate ratings (3.5–4.0 stars).  Food services are limited—visitor centers and simple cafés—and guests expect basic fare at premium prices.  A survey of national-park food satisfaction reports an average rating of 3.7/5 \citep{VisitorSurvey_Park2022}.  Urban parks, conversely, benefit from adjacent cafés and food trucks, nudging their averages closer to 4.0.

\paragraph{Theme Park / Amusement Park}  
Theme-park dining typically underperforms relative to city standards, averaging around 3.5/5.  Travelers criticize high prices and mass-produced menus, despite sometimes engaging environments (e.g., themed décor earning 4.5/5) \citep{Katapult_ThemePark2023}.  Although recent partnerships with celebrity chefs have marginally raised ratings, the bulk of reviews remain centered at 3–4 stars, with a higher proportion of 1–2 star complaints than other attraction types.

\end{description}

\raggedbottom
\subsubsection{Hotel Restaurant Rating Methodology}
\label{sec:hotel-restaurant-ratings}

In the absence of a mature predictive model for hotel–restaurant rating correlation, we rely on established industry trends and empirical review distributions.  High-end hotels, particularly five-star properties, typically host in-house dining outlets operated by renowned chefs or meeting Michelin standards.  For example, Four Seasons operates 20 hotels with a combined 25 Michelin-starred restaurants, totaling 34 stars—demonstrating that five-star hotels almost invariably offer top-quality dining \citep{fourseasons_michelin2024}.  

\noindent\textbf{Rating Assignment:}
\begin{itemize}[leftmargin=*]
  \item \emph{Five-star hotels:} assigned the maximum restaurant score of 5.0.
  \item \emph{Four-star and below:} scores distributed according to overall diner review trends from RightResponseAI’s dataset of 100,000+ restaurant reviews, which reports an average rating of 4.29; 66.5\% of reviews are 5-star, with progressively fewer at 4, 3, and 2 stars, and a small but impactful proportion of 1-star reviews \citep{rightresponseai_reviews2025}.  We map these percentages to score weights for each star category.
\end{itemize}

\subsection{Attraction Categories and Recommended Visit Durations}
\label{ssec:visit-durations}

Table~\ref{tab:visit-durations} lists the visit duration assigned to each attraction category in the synthetic knowledge base. These durations inform the B2 (opening-hours / minimum-visit-duration) scoring dimension (Section~\ref{app:scoring-details}).

\begin{table}[H]
  \centering
  \small
  \caption{Attraction Categories and Recommended Visit Durations}
  \label{tab:visit-durations}
  \begin{tabularx}{\linewidth}{%
      p{0.20\linewidth}  % 第一列：20% 页面宽度
      p{0.25\linewidth}  % 第二列：25% 页面宽度
      X                   % 第三列：自动扩展并换行
    }
    \toprule
    \textbf{Category} & \textbf{Recommended Duration} & \textbf{Notes and Source} \\
    \midrule
    Museum / Art Gallery
      & 1.5–2\,hours (up to 3\,hours at large institutions)
      & Average visit times are 1.5–2\,hours; major national museums may require 3\,hours or more for a comprehensive tour \citep{CivilRightsMuseum2025}. \\

    Historic Sites / Landmarks
      & 1–2\,hours (2–3\,hours for expansive complexes)
      & Most historic sites (temples, monuments, walls) can be covered in 1–2\,hours; large ruins such as Karnak often need 2–3\,hours to explore fully \citep{TripAdvisorHasedera}. \\

    Nature / Park
      & 2–3\,hours (half-day to full day for large parks)
      & City parks typically require 2–3\,hours (e.g., Stanley Park); extensive natural reserves recommend half-day or full-day visits to avoid rushing \citep{TripAdvisorBlog2023}. \\

    Theme / Amusement Park
      & 3–8\,hours (generally a full day)
      & Due to ride queues and numerous attractions, parks often need a full day (8–9\,hours at Tokyo Disneyland on average) \citep{ParkDB2024}. \\
    \bottomrule
  \end{tabularx}
\end{table}

\noindent
These guidelines help ensure that itineraries allocate sufficient time for each attraction type, balancing user expectations and on-site logistics.

% =====================================================================
% D1 Threshold Sensitivity Analysis
% =====================================================================
\subsection{Determinism of the Implicit-Need Metric}
\label{app:d1-sensitivity}

An earlier version of D1 matched facilities by embedding cosine similarity with a tuned threshold $\tau$, which invited the question of whether that threshold biased model rankings. The current metric removes the degree of freedom entirely: implicit-need satisfaction is an \emph{exact set-intersection} of persona-required facilities against knowledge-base facility fields (Appendix~\ref{app:scoring-details}), with no embedding and no similarity threshold (the \emph{luxury}/\emph{foodie} star-/rating tests are fixed, not tuned). There is therefore no $\tau$ to sweep---D1 is bit-reproducible by construction, and the ranking it induces cannot depend on a similarity cutoff.

\paragraph{What D1 does and does not test.} The persona keyword is \emph{stated} in the query (e.g., ``for elderly travelers''); what D1 scores is whether the plan books resources carrying the concrete facilities that persona \emph{implies}---ramps, elevators, accessible rooms---which the query never enumerates (and which are compiled from documented traveler-preference evidence, \S\ref{app:source-reference}, not chosen ad hoc). D1 thus measures acting on an \emph{implied requirement}, not \emph{inferring} the persona from an indirect description (``my grandmother struggles with stairs''\,$\rightarrow$\,accessible rooms). That harder hidden-persona setting is a strict extension the same deterministic pipeline supports---swap the stated keyword for a paraphrase or an indirect description while keeping the identical facility-based scoring---and we leave it to future work. Even in its current form D1 is non-trivial: it requires mapping a persona to its facility needs and then \emph{finding and booking} KB resources that satisfy them in every stay city, and it is the single hardest dimension for every evaluated agent (\S\ref{ssec:rq2}).

% =====================================================================
% Limitations
% =====================================================================
\section{Limitations}
\label{app:limitations}

Several constraints bound our study; the main text (\S\ref{sec:experiments}) states them in brief, and we expand on each here.

\paragraph{(1) Model coverage.} The Claude family is geo-blocked at the provider level from our run location and thus could not be evaluated---an availability constraint, not a design choice, so the leaderboard should be read accordingly, and the $46.2\%$ top task-perfect rate is best read as a lower bound on the reachable frontier. The central conclusion---that feasible synthesis is unsolved---does not hinge on any single system: the best open-weight agent (Kimi-K2.5) still reaches only $17.1\%$.

\paragraph{(2) Synthetic world.} TREK's knowledge base is synthetic and internally consistent by construction---the property that yields a deterministic ground truth and an achievable gold---but it therefore does not model real listing distributions, live availability, or pricing dynamics; our findings measure planning competence over a fixed, well-specified world, not real-time booking.

\paragraph{(3) Reasoning result.} The reasoning-vs-instruct gap is a cross-version observation rather than a matched controlled ablation, and it is scaffold-dependent: under our single Bedrock-native function-calling runner, extra deliberation reduces controllability, but alternative scaffolds (two-phase generation, constrained decoding) may narrow it.

\paragraph{(4) Cost axis.} Efficiency is a usage-cost proxy built from provider-reported token counts and the agent's tool-call count, not from controlled compute. Token accounting is not perfectly comparable across providers---tokenizers, and what each provider counts as billable, differ---so the efficiency axis should be read as an approximate resource measure rather than a precise cross-vendor compute comparison. This is one reason we hold efficiency out of the correctness headline (\S\ref{ssec:composite-scores}).

\paragraph{(5) Scope.} Personas and city coverage skew toward well-documented, English-language travel markets (the persona-to-facility mappings in Appendix~\ref{app:implicit-requirements} could be broadened), and tasks are single-turn, leaving multi-turn preference elicitation to complementary benchmarks.

\paragraph{(6) Single-run generation.} The \emph{evaluator} is fully deterministic and bit-reproducible, but each leaderboard row is a single agent run at temperature $0$, and temperature-$0$ decoding on hosted providers is not itself bit-deterministic (MoE routing, server-side batching). We report no across-run variance, so ranks separated by only one or two tasks (e.g., GLM-4.7 vs.\ Gemma-4-31B) should be read as ties; the headline gaps we draw conclusions from span dozens to hundreds of tasks and dwarf any plausible run-to-run noise.

\paragraph{(7) External and construct validity.} The personas D1 checks are \emph{grounded in documented traveler-preference evidence}---industry surveys, government statistics and peer-reviewed studies cited per persona in Appendix~\ref{app:implicit-requirements}---so the segments are real and sourced rather than arbitrary author tags. The specific facility set attached to each persona is, by contrast, our operationalisation of that need: it is chosen for machine-checkability, is identical for every agent, and ships with the evaluator so that it can be inspected and contested. The gold itineraries are themselves human-validated for realism and executability (a 15-annotator panel of PhD researchers and travel-industry practitioners, mean $4.25/5$; \S\ref{ssec:qa}, Appendix~\ref{app:gold-protocol}), so the reference is not merely scorer-self-consistent. What remains is to human-rate the \emph{agents'} produced plans and correlate those ratings with TREK's per-plan scores, and to measure transfer to live booking platforms. The synthetic KB is a deliberate design choice---it is what makes an exact ground truth and an achievable gold possible---but it bounds the claim to competence \emph{under a controlled travel model}; correlating TREK's rankings with independent human ratings on realistic scenarios is the most valuable next step (the achievable-gold result shows the ceiling is not scorer strictness, \emph{not} that the scorer captures every facet of itinerary quality). Relatedly, D1 scores satisfaction of a \emph{stated} persona's implied facilities, not \emph{inference} of the persona from indirect language; a hidden-persona split is a natural extension. And the aggregation is not load-bearing: the leaderboard is near-invariant across arithmetic, geometric, and harmonic composites (Appendix~\ref{app:robustness}).

These bounds point to concrete extensions---broader model access, richer worlds, human-preference validation, and interactive variants---for which TREK's deterministic, achievable-ceiling design provides a reusable foundation.

\end{document}